\documentclass{article}

\usepackage[dblblindworkshop, preprint]{neurips_2026}

\usepackage[utf8]{inputenc}
\usepackage[T1]{fontenc}
\usepackage{hyperref}
\usepackage{url}
\usepackage{booktabs}
\usepackage{amsfonts}
\usepackage{amsmath}
\usepackage{amssymb}
\usepackage{graphicx}
\usepackage{microtype}
\usepackage{xcolor}
\usepackage{cleveref}

\crefname{appendix}{Appendix}{Appendices}
\Crefname{appendix}{Appendix}{Appendices}

\graphicspath{{figs/}}
\title{Concept Bottleneck Models in the Wild: \\ Assessing Interpretability under Image Corruptions}
\workshoptitle{Trustworthy AI for Good @ NeurIPS 2026}

\author{%
  \textbf{Kexin Zhang}$^1$ \quad
  \textbf{Katharina Prasse}$^1$ \quad
  \textbf{Mishal Fatima}$^1$ \quad
  \textbf{Margret Keuper}$^{1,2}$ \\
  $^1$Chair of Machine Learning, Data and Web Science Group, University of Mannheim \\$^2$Max-Planck-Institute for Informatics, Saarland Informatics Campus \\
 \texttt{kexin.zhang@students.uni-mannheim.de} \quad
}

\begin{document}
\maketitle

\begin{abstract}
Concept Bottleneck Models (CBMs) route every prediction through a layer of named, human-auditable concepts, which makes their decisions inspectable.
CBMs are mostly evaluated on clean benchmark images; we analyze how common image corruptions that mimic real sensor failures affect three CBMs on ImageNet and CUB.
Image corruptions reduce model accuracy by up to $66$ points on ImageNet and $73$ points on CUB, depending on corruption type and severity.
The damage shows mainly in the concept activations: at the harshest severity, the CLIP embedding keeps a cosine similarity of at least $0.59$ with its clean counterpart, yet the standardized concept activations are re-ranked so that the top concept changes in $82$--$91\%$ of images on ImageNet and $89$--$100\%$ on CUB on average. We find that the reordering is uneven across concepts. A per-concept shift score $d_k$ measures how far each concept moves under corruption. A linear head retrained on the most stable concepts beats the deployed head on corrupted images for all three models, by $0.7$--$1.3$ points on ImageNet and $1.8$--$9.5$ points on CUB (severity-3 mean over ten corruption types), while a random subset of the same size never does. Clean accuracy drops by at most $0.5$ points ($1.6$ for one model). Pruning to the most stable concepts is thus a first step towards corruption-robust CBMs. We release code and checkpoints at \url{https://github.com/kexinkk/cbm-in-the-wild}.

\end{abstract}

\section{Introduction}
\label{sec:intro}

When a model makes a decision that requires human review, a readable explanation of the decision is needed, which modern visual encoders cannot provide. Encoders such as CLIP deliver strong recognition from web-scale pretraining, but their predictions emerge from about 86 million image-encoder parameters that no human-scale analysis will trace end to end~\citep{radford_learning_2021}. Concept Bottleneck Models are inherently interpretable by learning a class mapping based on an intermediate layer of named, human-interpretable concepts~\citep{koh_concept_2020}.
Variants differ mainly in how the concept layer is built, either from language-derived concept sets in Label-free and Post-hoc CBMs~\citep{oikarinen_label-free_2023, yuksekgonul_post-hoc_2023} or from a bank of visual concepts clustered from crops of a few dozen training images per class in DCBM~\citep{prasse_dcbm_2025}. 
The concept bank is small enough for a person to audit.

Deployed systems receive corrupted inputs, such as sensor noise, blur, weather, compression artifacts, and dead pixels, and the accuracy of image classifiers is known to degrade under such shifts~\citep{hendrycks_benchmarking_2019, bhojanapalli2021understanding}. For a CBM, corruption affects not only the prediction but also its explanation: when the concept activations change, each decision is explained by concepts computed from corrupted evidence. This failure is hard to notice, because the model still returns a prediction with a concept explanation. Users who rely on the explanation may then act on a wrong decision without noticing. 

\begin{figure}[htbp]
    \centering
    \includegraphics[width=\linewidth]{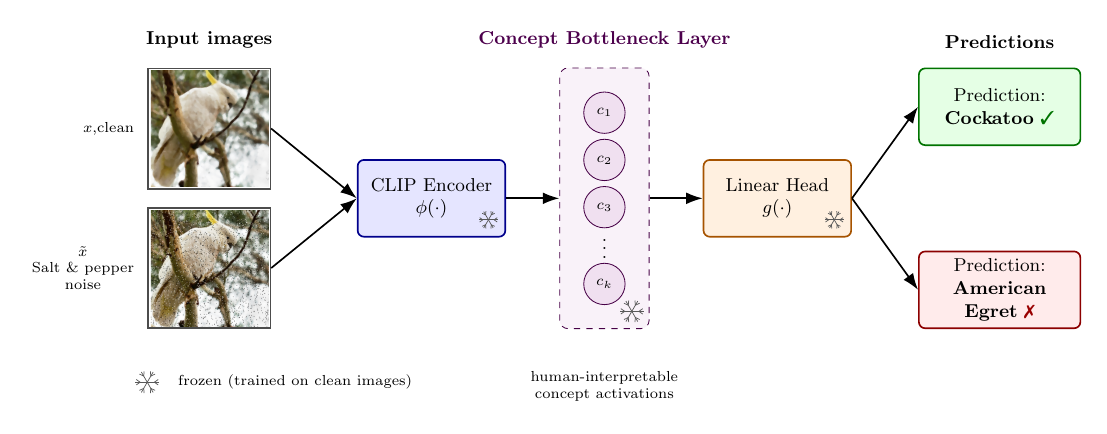}
    \caption{A deployed CBM routes every prediction through a bottleneck of concepts, and all three stages, the CLIP encoder, the concept bottleneck layer, and the linear head, are built or fit on clean images and stay frozen at test time (snowflakes). The clean image $x$ is correctly predicted as \emph{sulphur-crested cockatoo}. Salt-and-pepper noise at severity~3 leaves the bird visually distinct, yet the same frozen pipeline predicts \emph{American egret}.
    }
    \label{fig:fig1}
\end{figure}

This paper asks two questions for deployed CBMs. (1) Where does corruption do its damage as it propagates through the model, and what form does that damage take? (2) Does the damage have exploitable structure, so that a corruption-robust subset of concepts can be identified? We answer both in a setting that mimics deployment, where every component is fit on clean data and frozen, and corruption arrives only at test time through the input. All three CBMs share the same three-stage inference path once deployed, on which we run our analysis.
A frozen CLIP encoder embeds the input image, a frozen concept layer maps the embedding to $K$ concept activations, and a linear head maps the standardized scores to classes. Everything that builds the concept layer runs at training time on clean data only, and the head and the standardization statistics are fit on clean inputs as well, so corruption is a test-time-only shift that enters through the CLIP embedding of the corrupted input~(\Cref{fig:fig1}).


We first trace corruption through the pipeline by forwarding every image twice, once clean and once corrupted, and find that the concept activations change far more than the embedding: a moderate embedding shift becomes a re-ranking of the concept layer. A per-concept shift score $d_k$ then identifies a robust concept subset, which we compare against random subsets of the same size on three CBMs; for DCBM, we additionally separate its effect from crop resolution.

\paragraph{Contributions.}
(1) We show that image corruptions cost three deployed CBMs up to $66$ points of accuracy on ImageNet and up to $73$ on CUB, with Gaussian noise costing the most. The mildest corruption depends on dataset and model. On ImageNet, patch masking is the mildest corruption for all three CBMs, while on CUB, fog is the mildest one for DCBM and structured dropout for both LF-CBM and PCBM.
(2) We show that corruption damages a deployed CBM by \emph{reordering} its concept layer. The embedding direction maintains a cosine similarity $\geq 0.59$, yet the concept activation vector is reordered (\Cref{sec:propagation}). 
(3) We show that the reordering is uneven across concepts and that a per-concept shift score $d_k$ makes it actionable.
A head retrained on the most stable concepts beats the deployed full bank head on corrupted images on ImageNet and CUB, where a size-matched random subset never does (\Cref{sec:robust}).

\section{Setting: Deployed Pipeline, Data, and Protocol}
\label{sec:setting}
Once deployed for inference, the three models instantiate a common template in three stages described in ~\Cref{fig:fig1}. A frozen CLIP ViT-B/16 encoder ($D{=}512$, OpenAI release~\citep{radford_learning_2021} 
) embeds the full image, a frozen \emph{concept layer} maps the embedding to $K$ concept scores, and a linear head maps the standardized scores to classes. The models differ only in the concept layer. DCBM projects the embedding onto a bank of $K$ vectors clustered from CLIP embeddings of SAM2-segmented crops of clean training images~\citep{prasse_dcbm_2025, ravi_sam_2024}, while Label-free CBM (LF-CBM) and Post-hoc CBM (PCBM) read each score from a CLIP-text projection~\citep{oikarinen_label-free_2023, yuksekgonul_post-hoc_2023}. 
Across all three models, the head reads the per-concept activation, 
and DCBM alone inserts a softmax before the standardization. 
Everything that builds the concept layer is fit on clean data and never runs at test time, so the corrupted input reaches the model through a single channel, its CLIP image embedding.

\paragraph{Datasets and corruption suite.}
We study ImageNet-1000 and CUB-200-2011.
The three models differ in where their concepts come from:
DCBM builds its bank from the dataset's own training images ($K{=}2048$ concepts on ImageNet), while LF-CBM and PCBM build theirs from concept names through CLIP's text encoder ($K{=}4589$ on ImageNet).
Additionally, CIFAR-10 \citep{krizhevsky_cifar_2009} is used for the stage-wise propagation analysis~(\Cref{app:propagation},~\Cref{tab:app:grid_cifar10,tab:app:grid_lfcbm_cifar10,tab:app:grid_pcbm_cifar10}).
We corrupt inputs with ten corruption types, four natural corruptions from the standard benchmark of \citet{hendrycks_benchmarking_2019} and six custom sensor-level corruptions (construction and calibrated severity ladders in~\Cref{app:corruptions}  and~\Cref{fig:app:corruptiongrid}), at five severities calibrated by PSNR/SSIM. Severity-$5$ endpoints are calibrated to degrade accuracy clearly while staying above chance, so that damage remains attributable rather than saturated.


\paragraph{Protocol.}
We forward each image twice through the identical frozen pipeline, once clean and once corrupted, and read the paired, per-image representations. 
Accuracies are reported on the full test sets, the $50{,}000$-image ImageNet-1k validation set and the $5{,}794$-image CUB test set. 
For concept set analysis, we compare per-image concept ranks over paired test images of each (corruption, severity) cell.
For concept set pruning, we estimate $d_k$ on paired training images (the full training split on CUB, $50$ images per class on ImageNet), remove the concepts with the highest $d_k$, and retrain a linear head from scratch with each model's training recipe.
We evaluate a range of set sizes $K$ and select $K^\star$ by corrupted accuracy on the training split; all reported accuracies are on the disjoint test set.
\Cref{app:dk:estimates} compares estimating $d_k$ on training and test images and with one corruption type held out.
The evaluation is observational. Every step reads representations of a frozen pipeline; the model is neither trained nor adapted on corrupted inputs.
Seeding, determinism, and full compute details are in~\Cref{app:infra}; reproduction, the CLIP release, and baseline details are in~\Cref{app:repro}.

\section{Stage-Wise Propagation of Corruption}
\label{sec:propagation}

We follow corruption through the three stages of the deployed pipeline, the \emph{embedding} (CLIP), the \emph{concept activation} (frozen concept layer and standardization), and the \emph{prediction} (linear head), reading paired clean and corrupted representations as shown in ~\Cref{fig:fig1}. The central finding is general across the three models and is summarized in ~\Cref{tab:threefamily}. Under corruption, the embedding direction is largely preserved, while the concept activations the head reads are re-ranked. We carry the stage-wise mechanism in full on DCBM, while Label-free and Post-hoc CBM reproduce the reordering without a softmax, and their full per-corruption grids are in~\Cref{app:propagation}.
On CUB, the same signature holds for all three models at severity~$5$, an embedding cosine similarity of $0.73$, a top-$1$ concept that changes in $89$ to $100\%$ of images, and accuracy falling from $75$--$78\%$ clean to $16$--$29\%$ (\Cref{tab:threefamily}; per-corruption grids in ~\Cref{app:propagation}).

\begin{table}[t]
  \centering
  \caption{Stage-wise metrics at severity~$5$ on ImageNet (IMN) and CUB (ViT-B/16) for the three CBM models, each cell the mean over the ten corruption types; the clean top-1 row is the clean-reference accuracy. Reading down a column shows the amplification: a high embedding cosine similarity ($0.72$ on ImageNet and $0.73$ on CUB, identical across models by construction given the same encoder), a reordering of the standardized concept activations (the argmax-shift fraction is the fraction of images whose top-$1$ concept changes), and a collapsed prediction accuracy. Per-model, five-severity grids for both datasets are in ~\Cref{app:propagation}.}
  \label{tab:threefamily}
  \small
  \setlength{\tabcolsep}{4pt}
  \begin{tabular}{llcccccc}
    \toprule
    & & \multicolumn{2}{c}{DCBM} & \multicolumn{2}{c}{LF-CBM} & \multicolumn{2}{c}{PCBM} \\
    \cmidrule(lr){3-4} \cmidrule(lr){5-6} \cmidrule(lr){7-8}
    Stage & Metric (direction) & IMN & CUB & IMN & CUB & IMN & CUB \\
    \midrule
    Reference          & Clean top-1 $\uparrow$             & $0.70$ & $0.76$ & $0.75$ & $0.75$ & $0.72$ & $0.78$ \\
    \midrule
    Embedding          & Cosine similarity $\uparrow$      & $0.72$ & $0.73$ & $0.72$ & $0.73$ & $0.72$ & $0.73$ \\
    \midrule
    Concept activation & Argmax-shift fraction $\downarrow$ & $0.91$ & $1.00$ & $0.83$ & $0.94$ & $0.82$ & $0.89$ \\
                       & Top-5 Jaccard $\uparrow$           & $0.08$ & $0.00$ & $0.16$ & $0.10$ & $0.16$ & $0.16$ \\
                       & Spearman $\rho$ $\uparrow$          & $0.41$ & $0.20$ & $0.48$ & $0.44$ & $0.48$ & $0.49$ \\
    \midrule
    Prediction         & Top-1 accuracy $\uparrow$          & $0.27$ & $0.16$ & $0.35$ & $0.23$ & $0.32$ & $0.29$ \\
    \bottomrule
  \end{tabular}
\end{table}

\subsection{Stage 1: The CLIP Embedding}
Under corruption, the CLIP embedding is stable in direction. Measured as cosine similarity between paired clean and corrupted embeddings, it stays in $[0.59, 0.81]$ on ImageNet even at severity~$5$ (full per-corruption grid in ~\Cref{app:propagation}, ~\Cref{tab:app:grid_imagenet}), where images are visibly destroyed. CLIP is known to be comparatively robust to natural distribution shifts~\citep{radford_learning_2021, fang_data_2022}; we find that its embedding direction is also largely preserved under synthetic corruptions. The encoder is therefore not where the reordering is amplified. Even the corruptions that cost the model most of its accuracy leave the embedding at a cosine similarity $\geq 0.59$. The embedding does lose some information under severe corruption, but the main change can be observed in the concept activations and the label prediction.

\subsection{Stage 2: The Concept Activations}
The change in embedding direction leads to a near-total re-ranking of the concept layer. We read three rank statistics on the standardized activations $\tilde{\mathbf a}$ that the head consumes: the fraction of images whose top-$1$ concept changes, the Jaccard overlap of the top-$5$ concept set with its clean counterpart, and the Spearman rank correlation of the full $K$-vector. At severity~$5$ on ImageNet, across the three models, the top-1 concept changes in $82$--$91\%$ of images on average, with single corruptions leading to a change in $70$--$100\%$ of images (\Cref{tab:threefamily}). The reordering extends beyond the top-1 concept, with the top-$5$ concept overlap falling to $0.08$--$0.16$, so on average fewer than two of the five top concepts persist, and the full-vector correlation drops to $0.41$--$0.48$. On CUB the same three average statistics read $89$--$100\%$, $0.00$--$0.16$, and $0.20$--$0.49$, again with individual corruptions ranging further. Full details on per corruption reordering are in~\Cref{app:propagation}. 

\begin{figure}[t]
  \centering
  \includegraphics[width=0.8\textwidth]{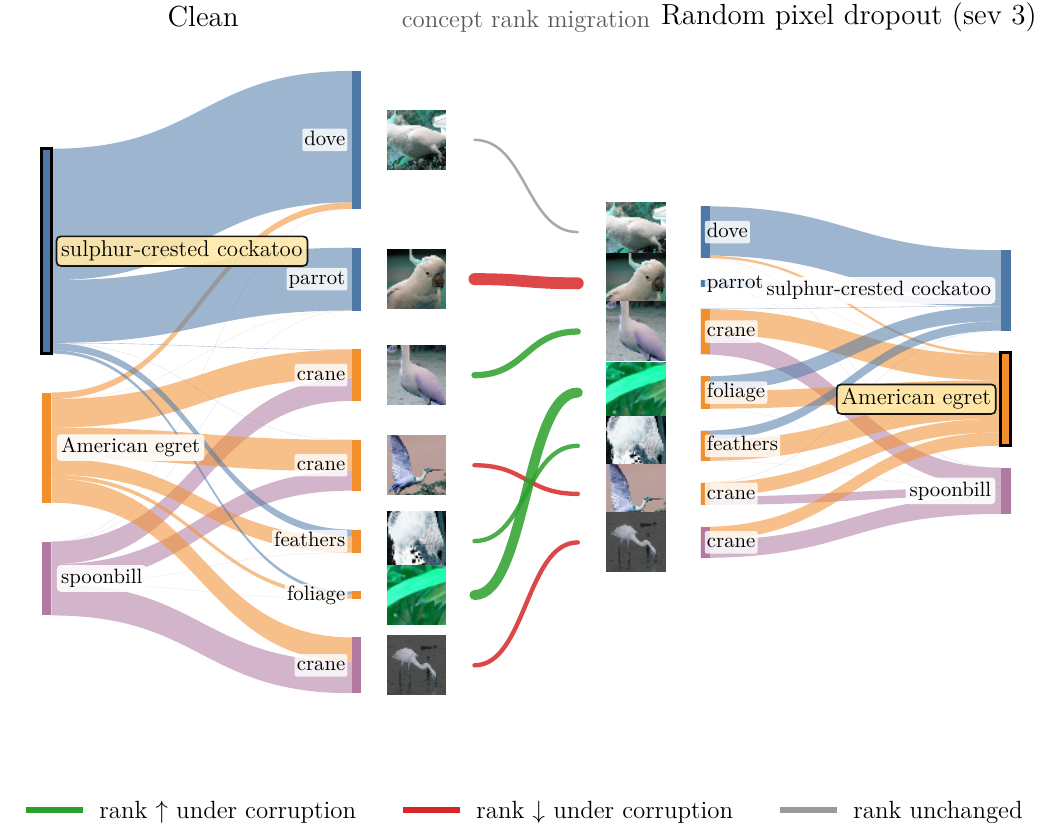} 
  \caption{Concept-to-prediction contribution flow for one ImageNet image, \emph{sulphur-crested cockatoo}, clean (left) versus random pixel dropout at severity~$3$ (right), the two panels mirrored about a rank-migration bridge. The prediction results are boxed and highlighted. A ribbon's width is a concept's positive contribution $\varphi_{k,c}=W_{c,k}\tilde a_k$ to a class; the bridge links each concept between its clean and corrupted contribution rank. The reordering drains contribution mass from the correct class and the prediction changes to \emph{American egret} (DCBM).}
  \label{fig:flow}
\end{figure}

The re-ranking is sharpest for DCBM. 
Its concept layer ends in a softmax over $K{=}2048$ near-equal scores, so the clean activations carry a per-concept spread of only $\sigma_k\sim10^{-5}$. The standardization divides by this spread, so activation shifts of order $10^{-5}$ become standardized changes of order one.
The text-projection models have no softmax bottleneck and reorder less (argmax shift $0.82$--$0.83$ against $0.91$ on ImageNet and $0.89$--$0.94$ against $1.00$ on CUB, ~\Cref{tab:threefamily}).
This amplified reordering is invisible to magnitude diagnostics. For DCBM, even at severity~$5$, the softmax fixes the activation $L_1$ norm at $1$, the relative $L_2$ change stays under $0.1\%$, and the activation entropy changes by less than $0.001$, so a norm- or entropy-based monitor reads the scrambled layer as untouched. 
~\Cref{fig:flow} shows the concept activation shift mechanism for one image. For the cockatoo, random pixel dropout moves the top clean concept's activation by $8\times10^{-5}$, from $6.6\times10^{-4}$ to $5.8\times10^{-4}$. Divided by its clean spread $\sigma_k=3\times10^{-5}$, this is a standardized drop from $+4.4$ to $+1.6$. Its contribution drains from the correct class, and the prediction turns over from \emph{sulphur-crested cockatoo} to \emph{American egret}. 

\subsection{Stage 3: The Prediction}
When the head reads the reordered activations, the top-$1$ accuracy falls to $5$--$54\%$ on ImageNet from a $70$--$76\%$ clean baseline, and to $3$--$43\%$ on CUB from $75$--$78\%$ clean (\Cref{tab:threefamily}).
On ImageNet at severity~5, the ten corruption types keep the identical worst-to-mildest order (\Cref{app:propagation}). Gaussian noise is harshest for every model, with top-1 accuracy $0.05$, $0.09$, and $0.08$ for DCBM, LF-CBM, and PCBM, and patch masking is mildest, with $0.47$, $0.54$, and $0.52$. 
The models differ in level, not in ordering: DCBM is $3$--$10$ points below the text-based models on every corruption type.
On CUB, Gaussian noise is again harshest for the three models at $3$ to $8\%$ top-1 and DCBM again sits lowest on every corruption type, though the gap to the text-projection models is no longer uniform and reaches $25$ points against PCBM on sparse occlusion.
The mildest corruption changes to fog for DCBM and structured dropout for LF-CBM and PCBM, so the harshest corruption is the same on both datasets, but the mildest is not~(\Cref{app:propagation} and \Cref{tab:app:grid_cub,tab:app:grid_lfcbm_cub,tab:app:grid_pcbm_cub}).

\section{The Per-Concept Shift and Concept Selection}
\label{sec:robust}
Whether the reordering can be exploited depends on how it is distributed across concepts. If corruption moved every concept by the same amount, the ranking would carry no per-concept structure, and no subset could be more robust than another.
We test this by measuring the shift, one concept at a time. For image $x$, let $\bar a_k(x)$ be concept $k$'s pre-softmax score. The per-concept shift $d_k$ is its mean absolute change, averaged over images (paired clean and corrupted) and over ten corruption types $F$ and five severities $S$,
\begin{equation}
  d_k = \frac{1}{|F|\,|S|}\sum_{f\in F}\sum_{s\in S}\mathbb{E}_x\bigl|\,\bar a^{\,f,s}_k(x)-\bar a^{\,\mathrm{clean}}_k(x)\,\bigr|,
  \label{eq:dk}
\end{equation}
so a low $d_k$ marks a concept whose score is stable under corruption. ~\Cref{app:dk} gives the estimator and its transfer to the other CBM models. The estimator runs on the paired training sets of ~\Cref{sec:setting}. 
We first show that $d_k$ is structured, and then that selecting concepts by $d_k$ outperforms size-matched random subsets for all three models on both datasets.

\subsection{The Distribution of the Per-Concept Shift}
\begin{figure}[htbp]
   \centering
  \includegraphics[width=0.9\textwidth]{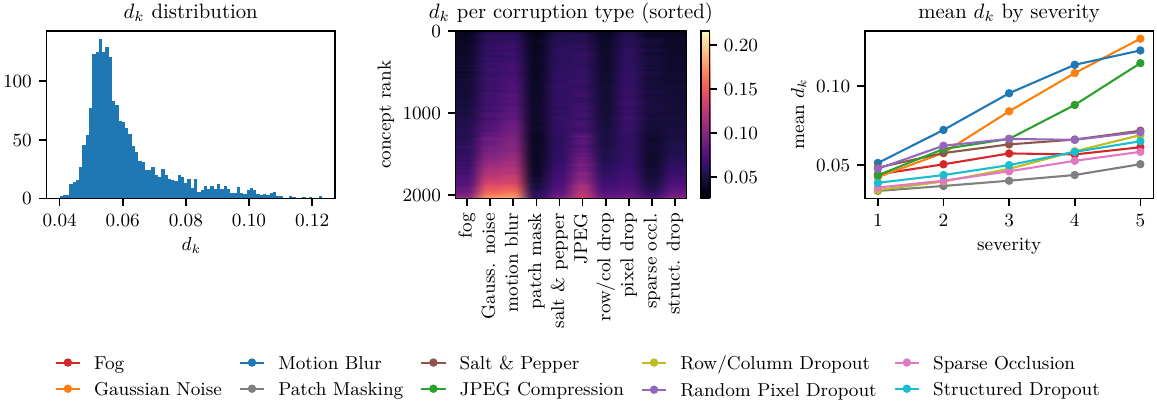}
  \caption{The per-concept shift $d_k$ for DCBM on the $K=2048$ ImageNet bank. Left: the distribution over concepts is unimodal with a mode near $0.05$ and a fragile right tail to about $0.12$. Middle: $d_k$ per corruption type; thus a single robust set can be found (concepts sorted by $d_k$). Right: $d_k$ increases with severity for every corruption type.}
  \label{fig:dkoverview}
\end{figure}

Across all three CBMs, the distribution of $d_k$ has the same shape, unimodal with a long right tail, so most concepts move little under corruption while a minority move several times as much. On DCBM's $2048$-concept bank, the mode sits near $0.05$, and the tail reaches about $0.12$, so the most fragile concepts move two to three times as far as the bulk (\Cref{fig:dkoverview}). Label-free CBM and Post-hoc CBM reproduce the same unimodal, long-tailed form on their $4589$-concept banks (\Cref{app:dk}, ~\Cref{fig:app:genericitydk}). For each model, the ranking is largely shared across corruption types, so one averaged $d_k$ defines a single robust set rather than ten sets that would each require knowing the corruption in advance (\Cref{fig:dkoverview}, middle). What the ranking looks like in image space is specific to DCBM, whose concepts are crops. Its \emph{robust} concepts, whose scores barely move, are large, sharp, roughly square crops of whole objects, while its \emph{fragile} concepts, whose scores move most, are small, thin, low-resolution slivers (medoid crops of each extreme in ~\Cref{app:dk}, ~\Cref{fig:app:robust,fig:app:fragile}). 

\subsection{Concept Selection}
\label{sec:selection}
We ask whether $d_k$ selects better concepts, not just fewer, and what pruning costs in clean accuracy.
At each size $K$, we compare against two references:
a size-matched \emph{random} subset of concepts and the deployed full model. 
The \emph{deployed} full bank head is the model as trained with the original recipe.
Every arm is a head retrained from scratch with the model's own recipe on the training split, the CLIP backbone and the concept vectors stay frozen, and no hyperparameter is re-tuned per $K$; the full bank retrain reproduces the deployed head (\Cref{sec:setting}, ~\Cref{app:protocol}), so a pruned head differs from it only in the pruning.

\begin{figure}[htbp]
  \centering
  \includegraphics[width=0.8\textwidth]{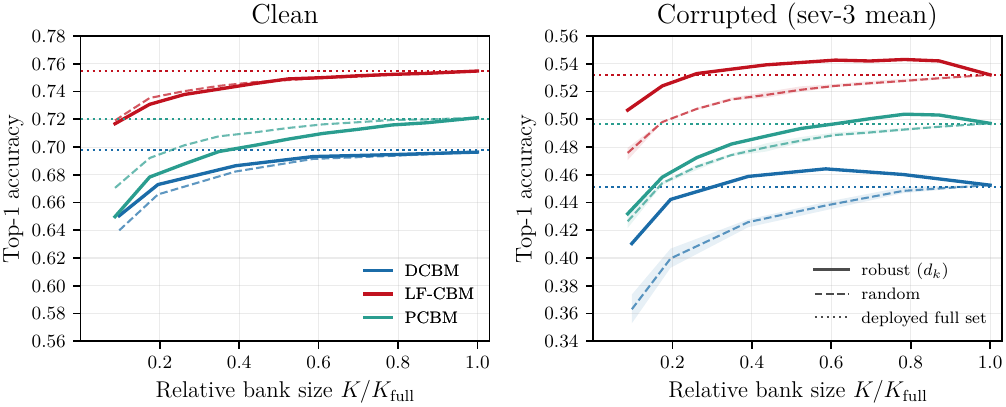}\\[2pt]
  \includegraphics[width=0.8\textwidth]{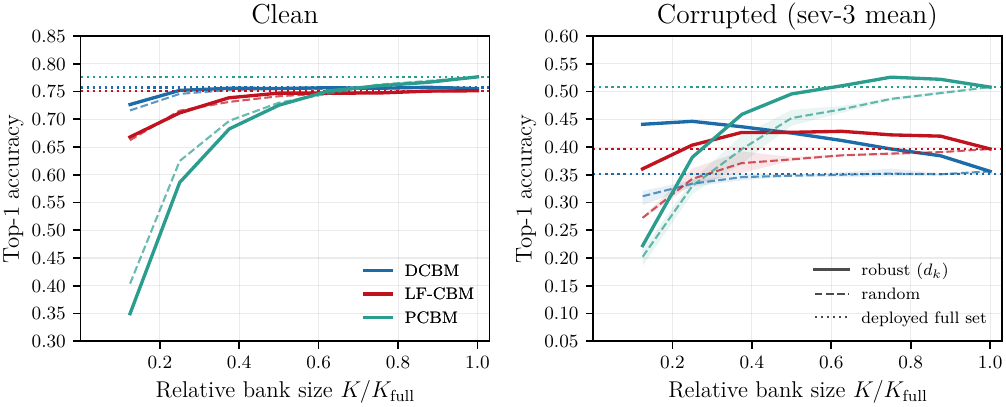}
  \caption{Concept selection on ImageNet (top) and CUB (bottom) for DCBM, LF-CBM and PCBM, each head retrained with the model's deployed recipe.
  Top-1 accuracy against the fraction of concepts kept $K/K_{\text{full}}$, clean (left) and corrupted (right, severity-3 mean over the ten corruption types).
  $d_k$ subset solid, random subset dashed with the min--max band over three draws, deployed full bank head dotted.
  The $d_k$ subset is above the random subset at every size and above the deployed head on corrupted images over the plateau of ~\Cref{tab:plateau}; on CUB, PCBM is ranked by the relative shift $\tilde d_k$ ($^{\ddagger}$). }
  \label{fig:genericity}
\end{figure}

We run this test on all three models on ImageNet and CUB, reusing the cached corrupted embeddings so no image model runs on corrupted data again. 
The shift $d_k$ is computed from each model's own concept score (\Cref{app:dk:def}).
Because PCBM's concept vectors have no common scale on CUB, we rank by the relative shift $\tilde d_k = d_k/\sigma_k$, the shift of the standardized activation the head reads, marked $^{\ddagger}$ (\Cref{app:dk}).
~\Cref{fig:genericity} and ~\Cref{tab:plateau} give the result.
At $K^\star$ (\Cref{tab:plateau}), the $d_k$ subset is above the random subset by $+2.6$, $+1.5$ and $+1.1$ points on ImageNet and $+11.2$, $+4.4$ and $+3.9$ on CUB; the margin is largest at smaller sizes, up to $+4.7$, $+3.1$ and $+1.1$ on ImageNet and $+13.0$, $+8.8$ and $+6.3$ on CUB (as reported in \Cref{fig:genericity,fig:app:pcbm,fig:app:lfcbm,fig:capacity}).
On ImageNet the plateau spans $K/K_{\text{full}}$ of $0.39$ to $0.78$ for DCBM, $0.26$ to $0.87$ for LF-CBM and $0.70$ to $0.87$ for PCBM, with a peak gain of $+1.3$, $+1.1$ and $+0.7$ points at a clean cost of $0.2$ to $0.5$ points. The gain comes from dropping the $22$ to $41\%$ most shift-sensitive concepts.
On CUB, with far fewer training images, the plateau is wider and the gain larger. DCBM is above its deployed head at every size of the grid, by $+9.0$ points at an eighth of its set ($3.1$ points clean cost) and $+9.5$ at a quarter ($0.5$), LF-CBM from a quarter of its set upward with $+3.2$ at $0.63$ of the set ($0.4$), and PCBM$^{\ddagger}$ from $0.63$ of its set upward with $+1.8$ at three quarters ($1.6$). 
A random subset of the same size is below the deployed head at every size, for every model, on both datasets, so the gain is selection and not size.
We show that only retraining reproduces the deployed head (\Cref{tab:app:protocols}), while estimating $d_k$ on training images, on test images, or with one corruption type held out gives comparable results (\Cref{tab:app:estimates}).

Zero-shot CLIP and a linear probe on the same embeddings (\Cref{tab:plateau}) serve as references without a concept bottleneck.
The probe is above every CBM on both axes, and on corrupted images zero-shot is above the deployed DCBM and PCBM on ImageNet and DCBM and LF-CBM on CUB.
On CUB, pruning lifts DCBM and LF-CBM above zero-shot on corrupted images, while on ImageNet it closes only part of the gap. Two DCBM-specific checks are in ~\Cref{app:casestudy}.
For its crop bank, pruning the large bank beats building a small one directly by $+5.6$ to $+12.9$ points of corrupted accuracy, and about half of DCBM's margin over random is crop resolution, free at build time, with a resolution-residualized ranking still beating random by $+3.1$ points at $K{=}200$.

\begin{table}[htbp]
  \centering
  \footnotesize
  \caption{Pruned concept set against the deployed full bank head.
  The $d_k$ subset is above the deployed head on corrupted images over the plateau of sizes, and a random subset of the same size never is. Deployed top-1 in \%; $\Delta$ the retrained head minus the deployed head in pp at $K^\star$, the size with the highest corrupted accuracy on the training set; corrupted is the severity-3 mean over the ten corruption types; head retrained with the model's recipe and reported on full test split. Zero-shot CLIP (ZS) and a linear probe (LP) on the same embeddings are references outside the bottleneck. $^{\ddagger}$ranked by the relative shift $\tilde d_k$ (\Cref{app:dk}).}
  \label{tab:plateau}
  \setlength{\tabcolsep}{4pt}
  \resizebox{\textwidth}{!}{%
  \begin{tabular}{llrlrrrr}
    \toprule
    & & & Deployed & Plateau $K$ & $K^\star$ & $\Delta$ $d_k$ & $\Delta$ random \\
    & Model & $K_{\text{full}}$ & clean / corr. & ($K/K_{\text{full}}$) & ($K/K_{\text{full}}$) & clean / corr. & clean / corr. \\
    \midrule
    ImageNet & DCBM   & 2048 & $69.8$ / $45.1$ & 800--1600 (0.39--0.78)  & 1200 (0.59) & $-0.5$ / $+1.3$ & $-0.6$ / $-1.3$ \\
             & LF-CBM & 4589 & $75.5$ / $53.2$ & 1200--4000 (0.26--0.87) &  3600 (0.78) & $-0.2$ / $+1.1$ & $-0.3$ / $-0.4$ \\
             & PCBM   & 4589 & $72.0$ / $49.7$ & 3200--4000 (0.70--0.87) &  3600 (0.78) & $-0.5$ / $+0.7$ & $-0.1$ / $-0.4$ \\
             & ZS & -- & $68.3$ / $50.7$ & & & & \\
             & LP & -- & $79.7$ / $56.7$ & & & &  \\
    \midrule
    CUB      & DCBM   & 1024 & $75.8$ / $35.1$ & 128--896 (0.12--0.88)   &  256 (0.25) & $-0.5$ / $+9.5$ & $-1.2$ / $-1.7$ \\
             & LF-CBM & 329  & $75.1$ / $39.7$ & 82--288 (0.25--0.88)    &  206 (0.63) & $-0.4$ / $+3.2$ & $-0.4$ / $-1.2$ \\
             & PCBM$^{\ddagger}$ & 112 & $77.7$ / $50.8$ & 70--98 (0.63--0.88) & 84 (0.75) & $-1.6$ / $+1.8$ & $-1.4$ / $-2.1$ \\
             & ZS & -- & $55.4$ / $41.0$ & & & &  \\
             & LP & -- & $81.4$ / $59.1$ & & & &  \\
    \bottomrule
  \end{tabular}}
\end{table}

\subsection{The Robust Subset: Sparse Interpretability}
The robust subset serves two purposes: diagnosis and sparsity.
From a single offline pass of paired forwards, $d_k$ identifies the concepts that change most under corruption. At $K^\star$, a head without them loses $0.2$--$0.5$ points of clean accuracy ($1.6$ for PCBM on CUB) and gains $0.7$--$9.5$ points on corrupted images, so the corrupted gain exceeds the clean cost by $0.2$--$9.0$ points (\Cref{tab:plateau}).
Pruning unstable concepts is a step towards robust accuracy, not a full repair: robust concepts still shift, since $d_k$ is a ranking and not a dichotomy, and part of the damage already occurs in the embedding.


\section{Related Work}
\label{sec:related}

\paragraph{Concept bottleneck models and the fragility of the concept layer.}
Concept Bottleneck Models route every prediction through named concepts~\citep{koh_concept_2020}, with recent variants attaching the bottleneck to a frozen encoder~\citep{prasse_dcbm_2025, oikarinen_label-free_2023, yuksekgonul_post-hoc_2023}. A line of work questions whether these concepts mean what they claim even on clean inputs. \citet{margeloiu_concept_2021} show that learned concepts need not correspond to their intended semantics, and \citet{mahinpei_promises_2021} and \citet{havasi_addressing_2022} trace leakage, in which the bottleneck encodes class information beyond the named concepts and the head comes to rely on it. \citet{debole_concept_2025} further find that vision-language concept supervision need not agree with expert labels, so concept accuracy does not guarantee concept quality. \citet{park_analysis_2025} corrupt concept annotations at training time and find that accuracy, interpretability, and intervention all degrade, with some concepts far more vulnerable than others. These studies show that concepts can be unreliable even on clean inputs. 
We study a distinct, test-time failure, in which corruption of the input reorders a clean-built layer with all parameters frozen.

\paragraph{Concept models under distribution shift and corruption.}
Robustness of concept models has been probed mostly through adversarial and shift lenses. \citet{sinha_understanding_2022} and \citet{li_evaluating_2025} show that small adversarial input changes flip concept predictions or interpretations while the model output stays stable. Closest to our setting, \citet{choi_adaptive_2025} argue that concept predictions are not inherently robust to distribution shift and adapt them with unlabeled target data, \citet{schmalwasser_faithfulness_2026} identify covariate shift and vision-language label noise as failure modes of Post-hoc CBM, \citet{eisenberg_spatially_2025} benchmark segment-level and text-derived CBMs on CIFAR-10-C, and \citet{uyterlinde_hyperbolic_2026} report improved corruption robustness from a hyperbolic concept geometry. Our study is deployment-facing, holding the clean-built bank frozen and reading the damage as a change of rank rather than repairing it with target data.

\paragraph{Concept selection and corruption recovery.}
\citet{kim_interpretability_2018} quantify a concept's importance through directional derivatives. For interventions, \citet{shin_closer_2023} show that which concepts one edits matters far more than how many, cutting task error by an order of magnitude at a fixed budget, the selection-over-quantity principle that our $d_k$ ranking carries over to corruption robustness. Our capacity-matched design, which fixes the bank size and varies only which concepts are kept, follows standard feature-selection practice~\citep{guyon_introduction_2003} and the build-large-then-prune ordering reported for trained networks~\citep{frankle_lottery_2019}. Corruption robustness is more often recovered at test time, by adapting normalization statistics to the corrupted input~\citep{schneider_improving_2020}, by entropy-based test-time adaptation~\citep{wang_tent_2021, niu_towards_2023}, or by weight-space ensembling~\citep{wortsman_robust_2022}. Our selection lever is instead fixed at build time. Estimating $d_k$ requires paired corrupted forwards once, offline, and nothing adapts at test time.

\paragraph{Backbone robustness and actionable interpretability.}
Our frozen backbone inherits CLIP's documented robustness to natural distribution shift~\citep{radford_learning_2021, fang_data_2022, tu_closer_2023}, and the corruption literature supplies the benchmark we build on~\citep{hendrycks_benchmarking_2019}. On the interpretability side, our approach answers longstanding calls to evaluate interpretability rigorously~\citep{doshi-velez_towards_2017} and by whether it enables concrete action~\citep{orgad_position_2026}. 

\section{Conclusion}
\label{sec:conclusion}

Across all three CBMs, corruption produces the same failure. It leaves the CLIP embedding direction largely preserved while almost completely reordering the concept activations. 
The failure is silent, since the pipeline keeps producing a concept explanation on corrupted evidence, so the auditability that motivates CBM adoption holds only as far as the concept layer is monitored. Here, rank rather than magnitude is the indicator of change according to our analysis. Whether such rank-based monitors can flag these failures in the field is a natural next step for trustworthy deployment.

Pruning by the per-concept shift $d_k$ yields a more robust concept subset.
A head retrained on these robust concepts beats the deployed full bank head on corrupted images for all three models on ImageNet and CUB, where a random subset of the same size fails.
The gain is $0.7$ to $1.3$ points on ImageNet and $1.8$ to $9.5$ on CUB.
We find that $d_k$ pruning is most useful for DCBM and LF-CBM, where the corrupted gain clearly exceeds the clean cost; for PCBM, the two are similar. $d_k$ pruning is thus a first step in making concept bottleneck models more robust to image corruptions. 
Future research could investigate how to prune the concept set without multiple retraining runs.

\paragraph{Limitations.}
Our claims rest on one backbone, CLIP ViT-B/16, and two datasets, ImageNet and CUB. 
The corruption suite is synthetic and covers ten types, so behavior under real sensor degradation is untested.
Our stage-wise statistics are correlational and do not decompose how much of the accuracy loss each stage causes.
$d_k$ is estimated offline from paired corrupted forwards. 
The proposed $d_k$ pruning requires multiple runs at different pruning capacities. 
We measure interpretability through the stability of concept activations and their ranking, a proxy that does not directly assess human interpretability or the faithfulness of the concept explanations.

\section*{Acknowledgements}
The authors acknowledge support by the state of Baden-Württemberg through bwHPC. Mishal Fatima and Margret Keuper acknowledge support by the DFG Research Unit 5336 - Learning to Sense (L2S). 

\bibliography{references}

\appendix
\crefalias{section}{appendix}
\crefalias{subsection}{appendix}

\newpage

\section{Infrastructure and Determinism}
\label{app:infra}
\paragraph{Seeds and determinism.}
We use two seeding regimes, matched to what each stage needs. The stage-wise forwards of ~\Cref{sec:propagation} and the per-concept shift $d_k$ of ~\Cref{sec:robust} are deterministic functions of a frozen pipeline, so they are computed once under a single fixed seed ($42$). Every result that involves a fitted component is reported over three seeds $\{0,1,2\}$. The capacity-matched selection test and its genericity replication draw the $K$-means \texttt{random\_state}, the random-subset arm, and the retrain arm from those seeds, so the reported margins carry the variance of both the selection draw and the head fit. A single entry point seeds Python, NumPy, and the PyTorch CPU and CUDA generators, sets \texttt{cudnn.deterministic}, disables the autotuner, and enables deterministic algorithms. \texttt{PYTHONHASHSEED} is pinned at launch, without which the hash-seeded per-image masks of the custom sensor corruptions would differ across runs while every other RNG stayed correctly seeded.

\paragraph{Environment and compute.}
All experiments ran under SLURM on NVIDIA H100 and A100 nodes, data on a Lustre parallel filesystem. The stack is Python~3.11 with PyTorch~2.1.2 on CUDA~12.1, pinned to the upstream DCBM requirements. SAM2 is used for concept-bank construction only. It runs once, offline, on the clean training subset and never touches a corrupted image. 

\section{Corruption Suite and Severity Calibration}
\label{app:corruptions}

\begin{figure}[htbp]
  \centering
  \includegraphics[width=\textwidth]{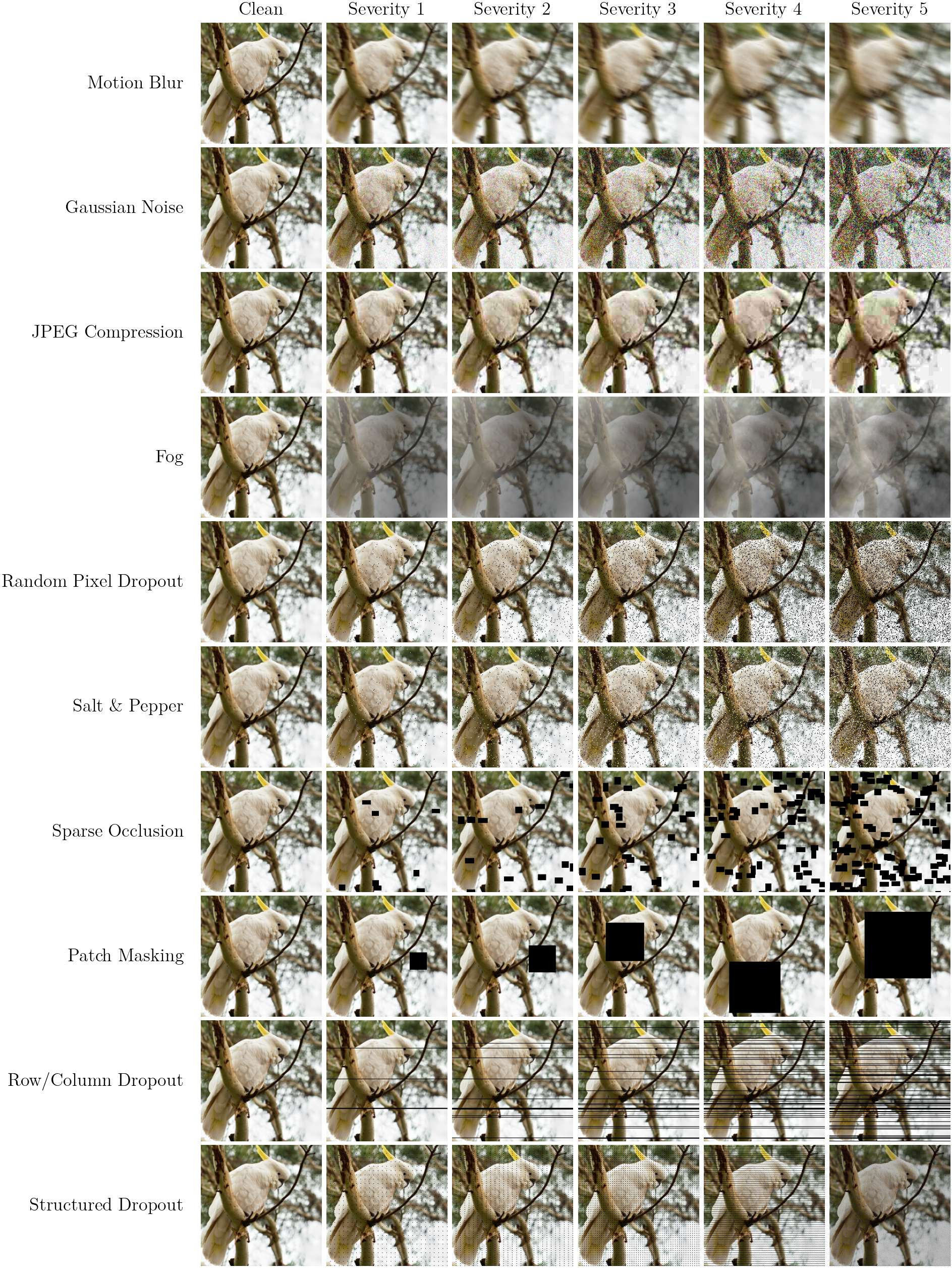}
  \caption{Severity monotonicity for all ten corruption types on one ImageNet image: clean reference and severities 1--5. The four natural corruptions (top rows) are the distributed ImageNet-C files; the six custom sensor corruptions are applied by the seedable transforms of ~\Cref{tab:app:sensorladder}.}
  \label{fig:app:corruptiongrid}
\end{figure}

The suite is ten corruption types per dataset, four natural corruptions taken verbatim from the distributed CIFAR-10-C and ImageNet-C files~\citep{hendrycks_benchmarking_2019}, namely motion blur, Gaussian noise, JPEG compression, and fog, covering the benchmark's blur, noise, digital, and weather categories at one representative each, and six custom sensor- and transmission-level corruptions the benchmark does not cover. Each custom corruption is a deterministic, seedable transform applied identically to all channels, mimicking a sensor rather than a photometric effect (\Cref{tab:app:sensorladder}). ~\Cref{fig:app:corruptiongrid} shows all ten corruptions across the clean-to-severity-$5$ range on one image.

\begin{table}[htbp]
  \centering
  \caption{The six custom sensor corruptions and their severity-level parameters (severities 1 to 5, canonical calibrated levels).}
  \label{tab:app:sensorladder}
  \begin{tabular}{lll}
    \toprule
    Corruption & Simulated failure & Severity parameter (1 to 5) \\
    \midrule
    Random pixel dropout & sparse dead pixels & drop prob.\ $.01/.03/.07/.13/.22$ \\
    Salt \& pepper & impulse noise, bit errors & pixel prob.\ $.01/.03/.07/.13/.22$ \\
    Sparse random occlusion & dust, droplets & area $.02/.05/.10/.18/.30$ \\
    Patch masking & large contiguous occluder & area $.02/.05/.10/.18/.30$ \\
    Row/column dropout & scan-line loss & line frac.\ $.02/.05/.10/.17/.25$ \\
    Structured pixel dropout & regular readout grid & drop frac.\ $\tfrac1{64}/\tfrac1{32}/\tfrac1{16}/\tfrac18/\tfrac14$ \\
    \bottomrule
  \end{tabular}
\end{table}

The custom severity levels cannot be copied from anywhere, so we calibrated them against image-similarity metrics before any model evaluation. For each (corruption, severity) cell we measured PSNR and SSIM between the clean and corrupted versions of $500$ class-stratified images per dataset, and required degradation to be monotone in severity and to fall inside a corridor anchored to the Hendrycks ladder (severity~1 near PSNR~$25$ with high SSIM, severity~5 near PSNR~$12$--$14$ and SSIM~$0.30$--$0.40$). ~\Cref{tab:app:calib_imagenet,tab:app:calib_cifar} give the full five-severity grids. Severity~5 is deliberately harsh but above chance, so that a stage-wise attribution is not confounded by stages that have all collapsed to random. One asymmetry is by design. Even after recalibration the custom corruptions at severity~5 are often harsher than the natural ones of the same level, so a fixed severity is comparable within a corruption type but does not denote equal difficulty across types. 

\begin{table}[htbp]
  \centering
  \small
  \caption{ImageNet severity calibration, PSNR (dB) and SSIM at each severity, mean over $500$ class-stratified images per cell. Degradation is monotone in severity for every corruption.}
  \label{tab:app:calib_imagenet}
  \begin{tabular}{lrrrrrrrrrr}
    \toprule
    & \multicolumn{2}{c}{S1} & \multicolumn{2}{c}{S2} & \multicolumn{2}{c}{S3} & \multicolumn{2}{c}{S4} & \multicolumn{2}{c}{S5} \\
    \cmidrule(lr){2-3}\cmidrule(lr){4-5}\cmidrule(lr){6-7}\cmidrule(lr){8-9}\cmidrule(lr){10-11}
    Corruption & P & S & P & S & P & S & P & S & P & S \\
    \midrule
    Motion blur            & 23.7 & 0.73 & 21.5 & 0.63 & 19.9 & 0.55 & 18.6 & 0.49 & 18.0 & 0.47 \\
    Gaussian noise         & 24.5 & 0.58 & 21.4 & 0.46 & 18.4 & 0.34 & 15.8 & 0.24 & 13.4 & 0.17 \\
    JPEG compression       & 29.0 & 0.85 & 28.0 & 0.83 & 27.4 & 0.81 & 26.0 & 0.76 & 24.7 & 0.71 \\
    Fog                    & 14.4 & 0.64 & 13.5 & 0.58 & 12.7 & 0.53 & 12.6 & 0.52 & 12.2 & 0.47 \\
    Random pixel dropout   & 26.2 & 0.85 & 21.5 & 0.66 & 17.8 & 0.49 & 15.1 & 0.38 & 12.8 & 0.30 \\
    Salt \& pepper         & 24.9 & 0.82 & 20.1 & 0.59 & 16.4 & 0.35 & 13.7 & 0.22 & 11.4 & 0.14 \\
    Sparse occlusion       & 23.9 & 0.97 & 19.7 & 0.92 & 16.7 & 0.85 & 14.3 & 0.74 & 12.3 & 0.61 \\
    Patch masking          & 24.6 & 0.97 & 20.4 & 0.94 & 16.8 & 0.88 & 14.3 & 0.80 & 11.8 & 0.67 \\
    Row/column dropout     & 23.8 & 0.92 & 19.4 & 0.80 & 16.3 & 0.65 & 13.9 & 0.49 & 12.3 & 0.36 \\
    Structured dropout     & 24.3 & 0.72 & 21.3 & 0.62 & 18.3 & 0.49 & 15.3 & 0.38 & 12.2 & 0.28 \\
    \bottomrule
  \end{tabular}
\end{table}

\begin{table}[htbp]
  \centering
  \small
  \caption{CIFAR-10 severity calibration, PSNR (dB) and SSIM at each severity, mean over $500$ class-stratified images per cell. Degradation is monotone in severity for every corruption.}
  \label{tab:app:calib_cifar}
  \begin{tabular}{lrrrrrrrrrr}
    \toprule
    & \multicolumn{2}{c}{S1} & \multicolumn{2}{c}{S2} & \multicolumn{2}{c}{S3} & \multicolumn{2}{c}{S4} & \multicolumn{2}{c}{S5} \\
    \cmidrule(lr){2-3}\cmidrule(lr){4-5}\cmidrule(lr){6-7}\cmidrule(lr){8-9}\cmidrule(lr){10-11}
    Corruption & P & S & P & S & P & S & P & S & P & S \\
    \midrule
    Motion blur            & 27.8 & 0.93 & 24.4 & 0.86 & 22.7 & 0.79 & 22.7 & 0.79 & 21.4 & 0.73 \\
    Gaussian noise         & 28.1 & 0.89 & 24.6 & 0.82 & 22.1 & 0.74 & 21.2 & 0.71 & 20.3 & 0.67 \\
    JPEG compression       & 31.2 & 0.95 & 29.4 & 0.94 & 28.9 & 0.93 & 28.3 & 0.92 & 27.6 & 0.91 \\
    Fog                    & 26.2 & 0.96 & 20.3 & 0.86 & 18.0 & 0.76 & 16.6 & 0.67 & 15.1 & 0.54 \\
    Random pixel dropout   & 26.1 & 0.90 & 21.1 & 0.78 & 17.3 & 0.62 & 14.6 & 0.50 & 12.3 & 0.38 \\
    Salt \& pepper         & 25.3 & 0.90 & 20.5 & 0.75 & 16.7 & 0.56 & 14.0 & 0.41 & 11.7 & 0.28 \\
    Sparse occlusion       & 23.5 & 0.88 & 19.2 & 0.76 & 16.3 & 0.61 & 14.0 & 0.48 & 12.0 & 0.37 \\
    Patch masking          & 22.8 & 0.92 & 19.9 & 0.88 & 16.4 & 0.79 & 13.3 & 0.63 & 11.1 & 0.45 \\
    Row/column dropout     & 21.0 & 0.88 & 18.1 & 0.78 & 16.1 & 0.70 & 13.8 & 0.56 & 11.8 & 0.41 \\
    Structured dropout     & 23.9 & 0.85 & 20.8 & 0.76 & 17.8 & 0.63 & 14.8 & 0.50 & 11.8 & 0.35 \\
    \bottomrule
  \end{tabular}
\end{table}

\section{Reproducibility}
\label{app:repro}
All three models use the same frozen CLIP ViT-B/16 backbone, the OpenAI CLIP reference implementation with its released ViT-B/16 weights and \texttt{clip.load} preprocessing (\url{https://github.com/openai/CLIP}). This appendix details DCBM's reproduction and bank construction, then records the Label-free CBM and Post-hoc CBM baselines.

\paragraph{DCBM reproduction and hyperparameter sweep.}
We reproduce the published DCBM baselines before any corruption work and select the operating point by a grid over bank size $K$, head $L_2$ weight $\lambda_1$, and learning rate (\Cref{tab:app:sweep}). Clean test accuracy rises monotonically with $K$ on ImageNet, from $60.2\%$ at $K{=}128$ to $69.77\%$ at $K{=}2048$, matching the authors' matched-bank $69.2\%$ to within $1\%$. A ten-fold heavier penalty $\lambda_1{=}10^{-3}$ collapses ImageNet accuracy (to $48.6\%$ at $K{=}2048$) while barely moving CIFAR-10, so we fix $\lambda_1{=}10^{-4}$ and learning rate $10^{-4}$ throughout. CIFAR-10 saturates near $94.8\%$ by $K{=}512$, consistent with its degenerate bank.

\begin{table}[htbp]
  \centering
  \caption{Bank-size and regularization sweep on CLIP ViT-B/16 (single seed, clean test top-1 accuracy). ImageNet is shown at the chosen $\lambda_1{=}10^{-4}$ and at the over-regularized $\lambda_1{=}10^{-3}$ (both learning rate $10^{-4}$); CIFAR-10 at $\lambda_1{=}10^{-4}$.}
  \label{tab:app:sweep}
  \begin{tabular}{lccc}
    \toprule
    & ImageNet & ImageNet & CIFAR-10 \\
    $K$ & $\lambda_1{=}10^{-4}$ & $\lambda_1{=}10^{-3}$ & $\lambda_1{=}10^{-4}$ \\
    \midrule
    $128$  & $0.602$ & $0.155$ & $0.945$ \\
    $256$  & $0.641$ & $0.241$ & $0.948$ \\
    $512$  & $0.670$ & $0.338$ & $0.948$ \\
    $1024$ & $0.685$ & $0.420$ & $0.948$ \\
    $2048$ & $0.698$ & $0.486$ & $0.948$ \\
    \bottomrule
  \end{tabular}
\end{table}

\paragraph{Medoid-versus-mean bank.}
We adopt the cluster medoid, the member closest to the coordinate-wise median, as the concept vector rather than the cluster mean upstream DCBM uses, so that every concept is a real segment crop with an exact image patch behind it. The choice costs no accuracy (\Cref{tab:app:bankablation}). This medoid-versus-mean comparison is a bank-construction control at $K{=}2048$, run under one fixed head so that only the concept-vector definition differs, and it is not the tuned deployed operating point, which is why its absolute level ($66.5\%$ for the medoid bank) sits below the deployed $69.77\%$ of ~\Cref{tab:app:sweep}. Read on that common footing the medoid bank reaches $66.5\%$ clean accuracy against $61.6\%$ for the mean bank at identical $K$. On CIFAR-10 the two coincide because the degenerate $K=215$ bank has too few crops per cluster to separate the median member from the mean.

\begin{table}[htbp]
  \centering
  \caption{Clean top-1 accuracy of the medoid and mean concept banks at identical $K$. The medoid bank matches or beats the mean bank while keeping every concept a real crop.}
  \label{tab:app:bankablation}
  \begin{tabular}{lccc}
    \toprule
    Dataset & $K$ & Medoid clean acc. & Mean clean acc. \\
    \midrule
    ImageNet & $2048$ & $0.665$ & $0.616$ \\
    CIFAR-10 & $215$  & $0.928$ & $0.928$ \\
    \bottomrule
  \end{tabular}
\end{table}

\paragraph{Label-free CBM and Post-hoc CBM baselines.}
On the same CLIP ViT-B/16 backbone, our reproduced clean top-1 accuracies are $75.5\%$ for Label-free CBM and $72.0\%$ for Post-hoc CBM (\Cref{tab:threefamily}). Both use the $K{=}4589$ concept bank released with Label-free CBM~\citep{oikarinen_label-free_2023}, which we reuse unchanged for Post-hoc CBM. Unlike DCBM's clustered crop bank, there is no bank-size hyperparameter to sweep here.

\paragraph{CUB models.}
On CUB-200-2011 ($200$ classes, $5{,}794$ test images; of the $5{,}994$ training images we hold out $600$ as a validation split and train on $5{,}394$) the deployed clean top-1 is $75.8\%$ for DCBM, with $K{=}1024$ concepts clustered from SAM2 crops of the training images, $75.1\%$ for LF-CBM, whose $329$ concepts are the curated CUB concept list of \citet{oikarinen_label-free_2023} ($369$ names) after its CLIP-similarity and interpretability cutoffs, and $77.7\%$ for PCBM, whose $K{=}112$ concept vectors are, one per CUB attribute of \citet{koh_concept_2020}, the normals of linear SVMs that separate clean training embeddings with the attribute from those without it.
Deployed is the best of three seeds on the clean test split, for every model on both datasets.

\section{Stage-Wise Propagation: Full Grids}
\label{app:propagation}
The severity-5 summary is given in the main text (\Cref{tab:threefamily}), which reads the same stage metrics on DCBM, Label-free CBM, and Post-hoc CBM (ImageNet and CUB, ViT-B/16) and shows the reordering signature holding for every family. For Post-hoc CBM, we use the multimodal (mmL) concept bank throughout, built from CLIP-text embeddings of concept names and the CAV bank built from the $112$ labelled attributes on CUB.

This appendix gives the full per-model, five-severity grids, in the order DCBM, Label-free CBM, Post-hoc CBM, each on ImageNet, then on CUB, then on CIFAR-10. ~\Cref{tab:app:grid_imagenet,tab:app:grid_cub,tab:app:grid_cifar10} give the complete DCBM stage-metric grids that ~\Cref{tab:threefamily} summarizes at severity~5. ~\Cref{tab:app:grid_lfcbm,tab:app:grid_lfcbm_cub,tab:app:grid_lfcbm_cifar10} give the matching grids for Label-free CBM, and ~\Cref{tab:app:grid_pcbm,tab:app:grid_pcbm_cub,tab:app:grid_pcbm_cifar10} for Post-hoc CBM. On CIFAR-10 all three models reproduce the reordering signature: from clean accuracies near $95\%$, the severity-5 mean argmax shift is $0.90$ for Label-free CBM, $0.93$ for Post-hoc CBM, and $0.97$ for DCBM, while accuracy falls to $0.54$--$0.56$ and the embedding cosine stays at $0.85$. Degradation is monotone in severity, and the per-corruption ordering is stable down the pipeline across all three models, on both datasets. On CUB all three models reproduce the signature at severity~5, with the embedding cosine at $0.73$, the mean argmax shift at $1.00$, $0.94$, and $0.89$ for DCBM, Label-free CBM, and Post-hoc CBM, and accuracy falling from $75$--$78\%$ clean to $0.16$, $0.23$, and $0.29$. Gaussian noise is the harshest corruption on both datasets for every model, while the mildest corruption differs, patch masking on ImageNet against fog (DCBM) or structured dropout (Label-free CBM, Post-hoc CBM) on CUB.

\begin{table}[htbp]
  \centering
  \small
  \caption{DCBM ImageNet full stage-metric grid (ViT-B/16), giving Stage-1 cosine similarity, the standardized Stage-2 argmax-shift fraction, top-5 Jaccard, and Spearman $\rho$, and Stage-3 top-1 accuracy, for every corruption type and severity. Clean-reference accuracy $69.77\%$ from deployed DCBM.}
  \label{tab:app:grid_imagenet}
  \begin{tabular}{llccccc}
    \toprule
    Corruption & S & cos\,$\uparrow$ & argmax\,$\downarrow$ & Jac\,$\uparrow$ & $\rho$\,$\uparrow$ & acc\,$\uparrow$ \\
    \midrule
    Clean & 0 & 1& 0&1 & 1& $0.70$ \\
    \midrule
    Motion blur            & 1 & 0.89 & 0.63 & 0.33 & 0.75 & 0.57 \\
                           & 2 & 0.84 & 0.77 & 0.20 & 0.64 & 0.48 \\
                           & 3 & 0.78 & 0.89 & 0.09 & 0.49 & 0.34 \\
                           & 4 & 0.72 & 0.95 & 0.04 & 0.36 & 0.20 \\
                           & 5 & 0.68 & 0.97 & 0.02 & 0.30 & 0.14 \\
    \midrule
    Gaussian noise         & 1 & 0.86 & 0.73 & 0.25 & 0.66 & 0.54 \\
                           & 2 & 0.82 & 0.81 & 0.17 & 0.57 & 0.45 \\
                           & 3 & 0.75 & 0.92 & 0.06 & 0.41 & 0.30 \\
                           & 4 & 0.66 & 0.98 & 0.02 & 0.26 & 0.15 \\
                           & 5 & 0.59 & 1.00 & 0.00 & 0.15 & 0.05 \\
    \midrule
    JPEG compression       & 1 & 0.88 & 0.61 & 0.36 & 0.77 & 0.58 \\
                           & 2 & 0.85 & 0.70 & 0.26 & 0.68 & 0.53 \\
                           & 3 & 0.84 & 0.75 & 0.22 & 0.64 & 0.49 \\
                           & 4 & 0.80 & 0.85 & 0.13 & 0.52 & 0.37 \\
                           & 5 & 0.74 & 0.93 & 0.06 & 0.39 & 0.24 \\
    \midrule
    Fog                    & 1 & 0.87 & 0.64 & 0.32 & 0.75 & 0.56 \\
                           & 2 & 0.85 & 0.70 & 0.27 & 0.70 & 0.52 \\
                           & 3 & 0.82 & 0.77 & 0.21 & 0.63 & 0.46 \\
                           & 4 & 0.80 & 0.82 & 0.17 & 0.58 & 0.40 \\
                           & 5 & 0.75 & 0.89 & 0.10 & 0.46 & 0.28 \\
    \midrule
    Random pixel dropout   & 1 & 0.87 & 0.58 & 0.38 & 0.78 & 0.58 \\
                           & 2 & 0.78 & 0.73 & 0.24 & 0.65 & 0.45 \\
                           & 3 & 0.73 & 0.83 & 0.15 & 0.55 & 0.35 \\
                           & 4 & 0.70 & 0.89 & 0.10 & 0.47 & 0.29 \\
                           & 5 & 0.67 & 0.94 & 0.05 & 0.39 & 0.23 \\
    \midrule
    Salt \& pepper         & 1 & 0.87 & 0.60 & 0.36 & 0.77 & 0.58 \\
                           & 2 & 0.80 & 0.74 & 0.24 & 0.65 & 0.46 \\
                           & 3 & 0.74 & 0.82 & 0.16 & 0.56 & 0.36 \\
                           & 4 & 0.70 & 0.88 & 0.10 & 0.47 & 0.28 \\
                           & 5 & 0.65 & 0.94 & 0.05 & 0.37 & 0.18 \\
    \midrule
    Sparse occlusion       & 1 & 0.89 & 0.55 & 0.40 & 0.77 & 0.63 \\
                           & 2 & 0.86 & 0.64 & 0.31 & 0.69 & 0.59 \\
                           & 3 & 0.82 & 0.73 & 0.22 & 0.63 & 0.53 \\
                           & 4 & 0.78 & 0.82 & 0.15 & 0.57 & 0.46 \\
                           & 5 & 0.75 & 0.89 & 0.09 & 0.50 & 0.38 \\
    \midrule
    Patch masking          & 1 & 0.91 & 0.48 & 0.46 & 0.81 & 0.65 \\
                           & 2 & 0.89 & 0.55 & 0.39 & 0.77 & 0.63 \\
                           & 3 & 0.86 & 0.62 & 0.33 & 0.72 & 0.60 \\
                           & 4 & 0.84 & 0.70 & 0.25 & 0.67 & 0.55 \\
                           & 5 & 0.81 & 0.80 & 0.16 & 0.59 & 0.47 \\
    \midrule
    Row/column dropout     & 1 & 0.91 & 0.45 & 0.49 & 0.84 & 0.65 \\
                           & 2 & 0.88 & 0.53 & 0.41 & 0.79 & 0.60 \\
                           & 3 & 0.85 & 0.65 & 0.31 & 0.71 & 0.53 \\
                           & 4 & 0.81 & 0.77 & 0.20 & 0.60 & 0.44 \\
                           & 5 & 0.77 & 0.87 & 0.11 & 0.50 & 0.35 \\
    \midrule
    Structured dropout     & 1 & 0.91 & 0.51 & 0.45 & 0.82 & 0.63 \\
                           & 2 & 0.88 & 0.59 & 0.36 & 0.73 & 0.60 \\
                           & 3 & 0.85 & 0.68 & 0.28 & 0.66 & 0.55 \\
                           & 4 & 0.81 & 0.80 & 0.17 & 0.55 & 0.46 \\
                           & 5 & 0.77 & 0.87 & 0.10 & 0.48 & 0.38 \\
    \bottomrule
  \end{tabular}
\end{table}

\begin{table}[htbp]
  \centering
  \small
  \caption{DCBM CUB full stage-metric grid (ViT-B/16, $K{=}1024$), giving Stage-1 cosine similarity, the standardized Stage-2 argmax-shift fraction, top-5 Jaccard, and Spearman $\rho$, and Stage-3 top-1 accuracy, for every corruption type and severity. Clean-reference accuracy $75.75\%$ from deployed DCBM.}
  \label{tab:app:grid_cub}
  \begin{tabular}{llccccc}
    \toprule
    Corruption & S & cos\,$\uparrow$ & argmax\,$\downarrow$ & Jac\,$\uparrow$ & $\rho$\,$\uparrow$ & acc\,$\uparrow$ \\
    \midrule
    Clean & 0 & 1 & 0 & 1 & 1 & 0.76 \\
    \midrule
    Motion blur            & 1 & 0.93 & 0.83 & 0.16 & 0.67 & 0.63 \\
                           & 2 & 0.88 & 0.95 & 0.05 & 0.49 & 0.48 \\
                           & 3 & 0.81 & 0.99 & 0.01 & 0.31 & 0.28 \\
                           & 4 & 0.73 & 1.00 & 0.00 & 0.18 & 0.14 \\
                           & 5 & 0.69 & 1.00 & 0.00 & 0.11 & 0.09 \\
    \midrule
    Gaussian noise         & 1 & 0.89 & 0.87 & 0.14 & 0.62 & 0.57 \\
                           & 2 & 0.85 & 0.96 & 0.06 & 0.46 & 0.44 \\
                           & 3 & 0.78 & 0.99 & 0.01 & 0.27 & 0.26 \\
                           & 4 & 0.69 & 1.00 & 0.00 & 0.12 & 0.11 \\
                           & 5 & 0.60 & 1.00 & 0.00 & 0.01 & 0.03 \\
    \midrule
    JPEG compression       & 1 & 0.86 & 0.95 & 0.07 & 0.48 & 0.49 \\
                           & 2 & 0.85 & 0.97 & 0.04 & 0.43 & 0.43 \\
                           & 3 & 0.84 & 0.98 & 0.03 & 0.38 & 0.38 \\
                           & 4 & 0.80 & 0.99 & 0.01 & 0.28 & 0.25 \\
                           & 5 & 0.75 & 1.00 & 0.00 & 0.18 & 0.14 \\
    \midrule
    Fog                    & 1 & 0.91 & 0.89 & 0.12 & 0.60 & 0.57 \\
                           & 2 & 0.89 & 0.94 & 0.06 & 0.51 & 0.49 \\
                           & 3 & 0.87 & 0.97 & 0.03 & 0.43 & 0.41 \\
                           & 4 & 0.86 & 0.97 & 0.03 & 0.41 & 0.38 \\
                           & 5 & 0.81 & 0.99 & 0.01 & 0.32 & 0.28 \\
    \midrule
    Random pixel dropout   & 1 & 0.89 & 0.87 & 0.14 & 0.64 & 0.58 \\
                           & 2 & 0.82 & 0.97 & 0.04 & 0.47 & 0.41 \\
                           & 3 & 0.76 & 0.99 & 0.01 & 0.34 & 0.30 \\
                           & 4 & 0.73 & 1.00 & 0.01 & 0.25 & 0.21 \\
                           & 5 & 0.69 & 1.00 & 0.00 & 0.18 & 0.14 \\
    \midrule
    Salt \& pepper         & 1 & 0.89 & 0.86 & 0.14 & 0.63 & 0.58 \\
                           & 2 & 0.83 & 0.95 & 0.06 & 0.49 & 0.46 \\
                           & 3 & 0.78 & 0.98 & 0.02 & 0.38 & 0.36 \\
                           & 4 & 0.74 & 0.99 & 0.01 & 0.28 & 0.27 \\
                           & 5 & 0.70 & 1.00 & 0.00 & 0.18 & 0.17 \\
    \midrule
    Sparse occlusion       & 1 & 0.86 & 0.96 & 0.04 & 0.47 & 0.47 \\
                           & 2 & 0.81 & 0.99 & 0.01 & 0.38 & 0.36 \\
                           & 3 & 0.78 & 0.99 & 0.01 & 0.33 & 0.29 \\
                           & 4 & 0.75 & 1.00 & 0.00 & 0.29 & 0.23 \\
                           & 5 & 0.73 & 1.00 & 0.00 & 0.24 & 0.17 \\
    \midrule
    Patch masking          & 1 & 0.89 & 0.90 & 0.10 & 0.55 & 0.56 \\
                           & 2 & 0.85 & 0.96 & 0.04 & 0.45 & 0.47 \\
                           & 3 & 0.82 & 0.98 & 0.02 & 0.37 & 0.38 \\
                           & 4 & 0.79 & 0.99 & 0.01 & 0.30 & 0.27 \\
                           & 5 & 0.75 & 1.00 & 0.00 & 0.22 & 0.16 \\
    \midrule
    Row/column dropout     & 1 & 0.90 & 0.85 & 0.14 & 0.65 & 0.60 \\
                           & 2 & 0.87 & 0.93 & 0.07 & 0.56 & 0.53 \\
                           & 3 & 0.84 & 0.97 & 0.03 & 0.46 & 0.42 \\
                           & 4 & 0.80 & 0.99 & 0.01 & 0.36 & 0.29 \\
                           & 5 & 0.77 & 1.00 & 0.00 & 0.28 & 0.21 \\
    \midrule
    Structured dropout     & 1 & 0.90 & 0.85 & 0.14 & 0.64 & 0.61 \\
                           & 2 & 0.88 & 0.95 & 0.05 & 0.52 & 0.52 \\
                           & 3 & 0.85 & 0.98 & 0.02 & 0.46 & 0.44 \\
                           & 4 & 0.81 & 1.00 & 0.00 & 0.35 & 0.31 \\
                           & 5 & 0.78 & 1.00 & 0.00 & 0.30 & 0.26 \\
    \bottomrule
  \end{tabular}
\end{table}

\begin{table}[htbp]
  \centering
  \small
  \caption{CIFAR-10 DCBM full stage-metric grid (ViT-B/16), giving Stage-1 cosine similarity, the standardized Stage-2 argmax-shift fraction, top-5 Jaccard, and Spearman $\rho$, and Stage-3 top-1 accuracy, for every corruption type and severity. Clean-reference accuracy $94.8\%$.}
  \label{tab:app:grid_cifar10}
  \begin{tabular}{llccccc}
    \toprule
    Corruption & S & cos\,$\uparrow$ & argmax\,$\downarrow$ & Jac\,$\uparrow$ & $\rho$\,$\uparrow$ & acc\,$\uparrow$ \\
    \midrule
    Motion blur            & 1 & 0.97 & 0.67 & 0.32 & 0.75 & 0.91 \\
                           & 2 & 0.94 & 0.82 & 0.18 & 0.58 & 0.86 \\
                           & 3 & 0.92 & 0.89 & 0.12 & 0.45 & 0.79 \\
                           & 4 & 0.92 & 0.89 & 0.12 & 0.45 & 0.79 \\
                           & 5 & 0.89 & 0.93 & 0.08 & 0.36 & 0.72 \\
    \midrule
    Gaussian noise         & 1 & 0.88 & 0.92 & 0.10 & 0.41 & 0.81 \\
                           & 2 & 0.86 & 0.95 & 0.06 & 0.29 & 0.67 \\
                           & 3 & 0.84 & 0.97 & 0.04 & 0.21 & 0.54 \\
                           & 4 & 0.83 & 0.98 & 0.03 & 0.18 & 0.46 \\
                           & 5 & 0.82 & 0.99 & 0.03 & 0.15 & 0.40 \\
    \midrule
    JPEG compression       & 1 & 0.92 & 0.89 & 0.13 & 0.51 & 0.84 \\
                           & 2 & 0.89 & 0.93 & 0.09 & 0.42 & 0.77 \\
                           & 3 & 0.89 & 0.94 & 0.08 & 0.40 & 0.75 \\
                           & 4 & 0.88 & 0.95 & 0.07 & 0.37 & 0.71 \\
                           & 5 & 0.86 & 0.96 & 0.06 & 0.32 & 0.66 \\
    \midrule
    Fog                    & 1 & 0.99 & 0.43 & 0.53 & 0.91 & 0.95 \\
                           & 2 & 0.97 & 0.67 & 0.32 & 0.77 & 0.93 \\
                           & 3 & 0.95 & 0.78 & 0.22 & 0.66 & 0.91 \\
                           & 4 & 0.93 & 0.86 & 0.16 & 0.56 & 0.87 \\
                           & 5 & 0.89 & 0.94 & 0.08 & 0.41 & 0.74 \\
    \midrule
    Random pixel dropout   & 1 & 0.94 & 0.78 & 0.23 & 0.65 & 0.93 \\
                           & 2 & 0.91 & 0.90 & 0.12 & 0.49 & 0.88 \\
                           & 3 & 0.88 & 0.94 & 0.08 & 0.37 & 0.78 \\
                           & 4 & 0.85 & 0.97 & 0.05 & 0.26 & 0.61 \\
                           & 5 & 0.83 & 0.98 & 0.03 & 0.18 & 0.41 \\
    \midrule
    Salt \& pepper         & 1 & 0.94 & 0.78 & 0.23 & 0.65 & 0.93 \\
                           & 2 & 0.91 & 0.90 & 0.12 & 0.51 & 0.87 \\
                           & 3 & 0.88 & 0.94 & 0.08 & 0.40 & 0.78 \\
                           & 4 & 0.85 & 0.96 & 0.05 & 0.28 & 0.64 \\
                           & 5 & 0.82 & 0.98 & 0.03 & 0.18 & 0.44 \\
    \midrule
    Sparse occlusion       & 1 & 0.94 & 0.81 & 0.20 & 0.59 & 0.93 \\
                           & 2 & 0.92 & 0.90 & 0.12 & 0.47 & 0.90 \\
                           & 3 & 0.89 & 0.94 & 0.08 & 0.35 & 0.83 \\
                           & 4 & 0.86 & 0.97 & 0.05 & 0.27 & 0.71 \\
                           & 5 & 0.84 & 0.98 & 0.04 & 0.20 & 0.56 \\
    \midrule
    Patch masking          & 1 & 0.96 & 0.79 & 0.22 & 0.63 & 0.93 \\
                           & 2 & 0.94 & 0.88 & 0.16 & 0.54 & 0.91 \\
                           & 3 & 0.92 & 0.93 & 0.10 & 0.45 & 0.86 \\
                           & 4 & 0.90 & 0.97 & 0.06 & 0.35 & 0.77 \\
                           & 5 & 0.86 & 0.98 & 0.04 & 0.25 & 0.62 \\
    \midrule
    Row/column dropout     & 1 & 0.96 & 0.75 & 0.25 & 0.68 & 0.93 \\
                           & 2 & 0.93 & 0.83 & 0.18 & 0.59 & 0.90 \\
                           & 3 & 0.91 & 0.88 & 0.15 & 0.51 & 0.87 \\
                           & 4 & 0.88 & 0.93 & 0.09 & 0.40 & 0.78 \\
                           & 5 & 0.84 & 0.97 & 0.05 & 0.27 & 0.64 \\
    \midrule
    Structured dropout     & 1 & 0.93 & 0.85 & 0.16 & 0.54 & 0.92 \\
                           & 2 & 0.91 & 0.91 & 0.11 & 0.44 & 0.88 \\
                           & 3 & 0.89 & 0.94 & 0.08 & 0.36 & 0.73 \\
                           & 4 & 0.85 & 0.98 & 0.04 & 0.23 & 0.54 \\
                           & 5 & 0.81 & 0.99 & 0.02 & 0.11 & 0.23 \\
    \bottomrule
  \end{tabular}
\end{table}

\begin{table}[htbp]
  \centering
  \small
  \caption{Label-free CBM ImageNet full stage-metric grid (ViT-B/16), giving Stage-1 cosine similarity, the standardized Stage-2 argmax-shift fraction, top-5 Jaccard, and Spearman $\rho$, and Stage-3 top-1 accuracy, for every corruption type and severity. Clean-reference accuracy $75.5\%$.}
  \label{tab:app:grid_lfcbm}
  \begin{tabular}{llccccc}
    \toprule
    Corruption & S & cos\,$\uparrow$ & argmax\,$\downarrow$ & Jac\,$\uparrow$ & $\rho$\,$\uparrow$ & acc\,$\uparrow$ \\
    \midrule
    Motion blur            & 1 & 0.89 & 0.55 & 0.41 & 0.80 & 0.65 \\
                           & 2 & 0.84 & 0.65 & 0.32 & 0.72 & 0.58 \\
                           & 3 & 0.78 & 0.78 & 0.21 & 0.58 & 0.44 \\
                           & 4 & 0.72 & 0.87 & 0.13 & 0.44 & 0.28 \\
                           & 5 & 0.68 & 0.91 & 0.09 & 0.37 & 0.20 \\
    \midrule
    Gaussian noise         & 1 & 0.86 & 0.58 & 0.38 & 0.78 & 0.62 \\
                           & 2 & 0.82 & 0.66 & 0.31 & 0.70 & 0.54 \\
                           & 3 & 0.75 & 0.78 & 0.20 & 0.56 & 0.40 \\
                           & 4 & 0.66 & 0.89 & 0.10 & 0.38 & 0.23 \\
                           & 5 & 0.59 & 0.96 & 0.04 & 0.22 & 0.09 \\
    \midrule
    JPEG compression       & 1 & 0.88 & 0.55 & 0.42 & 0.81 & 0.65 \\
                           & 2 & 0.85 & 0.60 & 0.37 & 0.76 & 0.61 \\
                           & 3 & 0.84 & 0.63 & 0.34 & 0.74 & 0.58 \\
                           & 4 & 0.80 & 0.72 & 0.26 & 0.66 & 0.47 \\
                           & 5 & 0.74 & 0.81 & 0.18 & 0.54 & 0.34 \\
    \midrule
    Fog                    & 1 & 0.87 & 0.58 & 0.38 & 0.78 & 0.64 \\
                           & 2 & 0.85 & 0.62 & 0.35 & 0.74 & 0.61 \\
                           & 3 & 0.82 & 0.68 & 0.29 & 0.68 & 0.55 \\
                           & 4 & 0.80 & 0.74 & 0.24 & 0.62 & 0.49 \\
                           & 5 & 0.75 & 0.84 & 0.15 & 0.50 & 0.36 \\
    \midrule
    Random pixel dropout   & 1 & 0.87 & 0.60 & 0.37 & 0.75 & 0.64 \\
                           & 2 & 0.78 & 0.73 & 0.25 & 0.62 & 0.53 \\
                           & 3 & 0.73 & 0.78 & 0.20 & 0.54 & 0.44 \\
                           & 4 & 0.70 & 0.81 & 0.17 & 0.48 & 0.37 \\
                           & 5 & 0.67 & 0.85 & 0.13 & 0.42 & 0.30 \\
    \midrule
    Salt \& pepper         & 1 & 0.87 & 0.61 & 0.36 & 0.76 & 0.65 \\
                           & 2 & 0.80 & 0.73 & 0.25 & 0.64 & 0.54 \\
                           & 3 & 0.74 & 0.80 & 0.19 & 0.56 & 0.44 \\
                           & 4 & 0.70 & 0.84 & 0.15 & 0.49 & 0.36 \\
                           & 5 & 0.65 & 0.88 & 0.11 & 0.40 & 0.26 \\
    \midrule
    Sparse occlusion       & 1 & 0.89 & 0.54 & 0.42 & 0.79 & 0.70 \\
                           & 2 & 0.86 & 0.63 & 0.34 & 0.72 & 0.66 \\
                           & 3 & 0.82 & 0.69 & 0.28 & 0.65 & 0.61 \\
                           & 4 & 0.78 & 0.75 & 0.23 & 0.59 & 0.55 \\
                           & 5 & 0.75 & 0.79 & 0.19 & 0.53 & 0.48 \\
    \midrule
    Patch masking          & 1 & 0.91 & 0.48 & 0.47 & 0.83 & 0.72 \\
                           & 2 & 0.89 & 0.54 & 0.41 & 0.78 & 0.69 \\
                           & 3 & 0.86 & 0.59 & 0.37 & 0.74 & 0.67 \\
                           & 4 & 0.84 & 0.65 & 0.32 & 0.70 & 0.63 \\
                           & 5 & 0.81 & 0.73 & 0.25 & 0.63 & 0.54 \\
    \midrule
    Row/column dropout     & 1 & 0.91 & 0.52 & 0.43 & 0.81 & 0.70 \\
                           & 2 & 0.88 & 0.62 & 0.35 & 0.75 & 0.64 \\
                           & 3 & 0.85 & 0.69 & 0.28 & 0.68 & 0.57 \\
                           & 4 & 0.81 & 0.75 & 0.23 & 0.61 & 0.50 \\
                           & 5 & 0.77 & 0.78 & 0.19 & 0.55 & 0.43 \\
    \midrule
    Structured dropout     & 1 & 0.91 & 0.50 & 0.45 & 0.83 & 0.69 \\
                           & 2 & 0.88 & 0.56 & 0.40 & 0.78 & 0.67 \\
                           & 3 & 0.85 & 0.61 & 0.35 & 0.74 & 0.63 \\
                           & 4 & 0.81 & 0.70 & 0.27 & 0.65 & 0.56 \\
                           & 5 & 0.77 & 0.76 & 0.22 & 0.60 & 0.48 \\
    \bottomrule
  \end{tabular}
\end{table}

\begin{table}[htbp]
  \centering
  \small
  \caption{Label-free CBM CUB full stage-metric grid (ViT-B/16, $K{=}329$), giving Stage-1 cosine similarity, the standardized Stage-2 argmax-shift fraction, top-5 Jaccard, and Spearman $\rho$, and Stage-3 top-1 accuracy, for every corruption type and severity. Clean-reference accuracy $75.13\%$.}
  \label{tab:app:grid_lfcbm_cub}
  \begin{tabular}{llccccc}
    \toprule
    Corruption & S & cos\,$\uparrow$ & argmax\,$\downarrow$ & Jac\,$\uparrow$ & $\rho$\,$\uparrow$ & acc\,$\uparrow$ \\
    \midrule
    Clean & 0 & 1 & 0 & 1 & 1 & 0.75 \\
    \midrule
    Motion blur            & 1 & 0.93 & 0.70 & 0.34 & 0.81 & 0.65 \\
                           & 2 & 0.88 & 0.82 & 0.24 & 0.72 & 0.53 \\
                           & 3 & 0.81 & 0.91 & 0.14 & 0.58 & 0.34 \\
                           & 4 & 0.73 & 0.95 & 0.09 & 0.45 & 0.21 \\
                           & 5 & 0.69 & 0.95 & 0.07 & 0.38 & 0.15 \\
    \midrule
    Gaussian noise         & 1 & 0.89 & 0.72 & 0.31 & 0.78 & 0.62 \\
                           & 2 & 0.85 & 0.79 & 0.24 & 0.70 & 0.51 \\
                           & 3 & 0.78 & 0.89 & 0.15 & 0.55 & 0.33 \\
                           & 4 & 0.69 & 0.94 & 0.08 & 0.37 & 0.18 \\
                           & 5 & 0.60 & 0.97 & 0.05 & 0.22 & 0.08 \\
    \midrule
    JPEG compression       & 1 & 0.86 & 0.76 & 0.28 & 0.73 & 0.55 \\
                           & 2 & 0.85 & 0.78 & 0.26 & 0.69 & 0.53 \\
                           & 3 & 0.84 & 0.80 & 0.24 & 0.66 & 0.48 \\
                           & 4 & 0.80 & 0.85 & 0.18 & 0.57 & 0.36 \\
                           & 5 & 0.75 & 0.91 & 0.12 & 0.46 & 0.22 \\
    \midrule
    Fog                    & 1 & 0.91 & 0.73 & 0.32 & 0.78 & 0.62 \\
                           & 2 & 0.89 & 0.78 & 0.27 & 0.73 & 0.56 \\
                           & 3 & 0.87 & 0.82 & 0.22 & 0.67 & 0.49 \\
                           & 4 & 0.86 & 0.82 & 0.21 & 0.64 & 0.46 \\
                           & 5 & 0.81 & 0.88 & 0.16 & 0.53 & 0.34 \\
    \midrule
    Random pixel dropout   & 1 & 0.89 & 0.80 & 0.24 & 0.74 & 0.53 \\
                           & 2 & 0.82 & 0.92 & 0.14 & 0.62 & 0.39 \\
                           & 3 & 0.76 & 0.96 & 0.10 & 0.56 & 0.31 \\
                           & 4 & 0.73 & 0.98 & 0.09 & 0.51 & 0.27 \\
                           & 5 & 0.69 & 0.98 & 0.08 & 0.44 & 0.20 \\
    \midrule
    Salt \& pepper         & 1 & 0.89 & 0.79 & 0.26 & 0.75 & 0.57 \\
                           & 2 & 0.83 & 0.88 & 0.17 & 0.65 & 0.44 \\
                           & 3 & 0.78 & 0.92 & 0.14 & 0.59 & 0.37 \\
                           & 4 & 0.74 & 0.92 & 0.14 & 0.53 & 0.31 \\
                           & 5 & 0.70 & 0.93 & 0.11 & 0.45 & 0.22 \\
    \midrule
    Sparse occlusion       & 1 & 0.86 & 0.86 & 0.19 & 0.70 & 0.51 \\
                           & 2 & 0.81 & 0.90 & 0.14 & 0.64 & 0.43 \\
                           & 3 & 0.78 & 0.92 & 0.13 & 0.59 & 0.38 \\
                           & 4 & 0.75 & 0.93 & 0.11 & 0.54 & 0.32 \\
                           & 5 & 0.73 & 0.94 & 0.10 & 0.49 & 0.25 \\
    \midrule
    Patch masking          & 1 & 0.89 & 0.84 & 0.20 & 0.72 & 0.53 \\
                           & 2 & 0.85 & 0.91 & 0.12 & 0.63 & 0.43 \\
                           & 3 & 0.82 & 0.95 & 0.09 & 0.56 & 0.33 \\
                           & 4 & 0.79 & 0.96 & 0.06 & 0.50 & 0.27 \\
                           & 5 & 0.75 & 0.97 & 0.05 & 0.41 & 0.17 \\
    \midrule
    Row/column dropout     & 1 & 0.90 & 0.83 & 0.21 & 0.73 & 0.57 \\
                           & 2 & 0.87 & 0.89 & 0.15 & 0.65 & 0.47 \\
                           & 3 & 0.84 & 0.92 & 0.11 & 0.59 & 0.38 \\
                           & 4 & 0.80 & 0.93 & 0.09 & 0.52 & 0.30 \\
                           & 5 & 0.77 & 0.94 & 0.09 & 0.47 & 0.26 \\
    \midrule
    Structured dropout     & 1 & 0.90 & 0.74 & 0.31 & 0.80 & 0.63 \\
                           & 2 & 0.88 & 0.77 & 0.27 & 0.76 & 0.59 \\
                           & 3 & 0.85 & 0.83 & 0.23 & 0.71 & 0.55 \\
                           & 4 & 0.81 & 0.88 & 0.17 & 0.63 & 0.46 \\
                           & 5 & 0.78 & 0.90 & 0.15 & 0.57 & 0.38 \\
    \bottomrule
  \end{tabular}
\end{table}

\begin{table}[htbp]
  \centering
  \small
  \caption{Label-free CBM CIFAR-10 full stage-metric grid (ViT-B/16, upstream CIFAR-10 concept set, $K{=}138$), giving Stage-1 cosine similarity, the standardized Stage-2 argmax-shift fraction, top-5 Jaccard, and Spearman $\rho$, and Stage-3 top-1 accuracy, for every corruption type and severity. Clean-reference accuracy $94.6\%$. The concept set differs from the ImageNet bank of ~\Cref{tab:app:grid_lfcbm}, so cells are comparable within this table, not against ImageNet.}
  \label{tab:app:grid_lfcbm_cifar10}
  \begin{tabular}{llccccc}
    \toprule
    Corruption & S & cos\,$\uparrow$ & argmax\,$\downarrow$ & Jac\,$\uparrow$ & $\rho$\,$\uparrow$ & acc\,$\uparrow$ \\
    \midrule
    Motion blur            & 1 & 0.97 & 0.46 & 0.55 & 0.89 & 0.91 \\
                           & 2 & 0.94 & 0.61 & 0.43 & 0.79 & 0.86 \\
                           & 3 & 0.92 & 0.72 & 0.33 & 0.70 & 0.80 \\
                           & 4 & 0.92 & 0.72 & 0.33 & 0.70 & 0.80 \\
                           & 5 & 0.89 & 0.80 & 0.27 & 0.61 & 0.73 \\
    \midrule
    Gaussian noise         & 1 & 0.88 & 0.83 & 0.25 & 0.64 & 0.82 \\
                           & 2 & 0.86 & 0.88 & 0.19 & 0.52 & 0.67 \\
                           & 3 & 0.84 & 0.91 & 0.15 & 0.41 & 0.53 \\
                           & 4 & 0.83 & 0.92 & 0.13 & 0.37 & 0.46 \\
                           & 5 & 0.82 & 0.93 & 0.11 & 0.32 & 0.40 \\
    \midrule
    JPEG compression       & 1 & 0.92 & 0.62 & 0.41 & 0.75 & 0.85 \\
                           & 2 & 0.89 & 0.71 & 0.33 & 0.66 & 0.78 \\
                           & 3 & 0.89 & 0.75 & 0.30 & 0.62 & 0.75 \\
                           & 4 & 0.88 & 0.78 & 0.26 & 0.58 & 0.72 \\
                           & 5 & 0.86 & 0.82 & 0.23 & 0.52 & 0.67 \\
    \midrule
    Fog                    & 1 & 0.99 & 0.25 & 0.74 & 0.97 & 0.94 \\
                           & 2 & 0.97 & 0.42 & 0.59 & 0.90 & 0.93 \\
                           & 3 & 0.95 & 0.53 & 0.49 & 0.84 & 0.91 \\
                           & 4 & 0.93 & 0.63 & 0.40 & 0.76 & 0.88 \\
                           & 5 & 0.89 & 0.80 & 0.26 & 0.59 & 0.76 \\
    \midrule
    Random pixel dropout   & 1 & 0.94 & 0.52 & 0.49 & 0.83 & 0.93 \\
                           & 2 & 0.91 & 0.72 & 0.32 & 0.69 & 0.89 \\
                           & 3 & 0.87 & 0.85 & 0.20 & 0.57 & 0.77 \\
                           & 4 & 0.85 & 0.92 & 0.14 & 0.47 & 0.56 \\
                           & 5 & 0.83 & 0.95 & 0.09 & 0.36 & 0.36 \\
    \midrule
    Salt \& pepper         & 1 & 0.94 & 0.52 & 0.50 & 0.83 & 0.93 \\
                           & 2 & 0.91 & 0.67 & 0.36 & 0.72 & 0.89 \\
                           & 3 & 0.88 & 0.81 & 0.24 & 0.58 & 0.78 \\
                           & 4 & 0.85 & 0.90 & 0.15 & 0.45 & 0.60 \\
                           & 5 & 0.82 & 0.96 & 0.08 & 0.32 & 0.38 \\
    \midrule
    Sparse occlusion       & 1 & 0.94 & 0.52 & 0.48 & 0.83 & 0.92 \\
                           & 2 & 0.92 & 0.65 & 0.38 & 0.74 & 0.90 \\
                           & 3 & 0.89 & 0.77 & 0.28 & 0.65 & 0.84 \\
                           & 4 & 0.86 & 0.87 & 0.20 & 0.54 & 0.70 \\
                           & 5 & 0.84 & 0.93 & 0.14 & 0.44 & 0.52 \\
    \midrule
    Patch masking          & 1 & 0.96 & 0.48 & 0.54 & 0.86 & 0.93 \\
                           & 2 & 0.94 & 0.55 & 0.47 & 0.81 & 0.92 \\
                           & 3 & 0.92 & 0.65 & 0.39 & 0.74 & 0.88 \\
                           & 4 & 0.90 & 0.75 & 0.29 & 0.63 & 0.80 \\
                           & 5 & 0.86 & 0.85 & 0.20 & 0.49 & 0.68 \\
    \midrule
    Row/column dropout     & 1 & 0.96 & 0.49 & 0.51 & 0.86 & 0.92 \\
                           & 2 & 0.93 & 0.62 & 0.42 & 0.79 & 0.91 \\
                           & 3 & 0.91 & 0.74 & 0.33 & 0.72 & 0.89 \\
                           & 4 & 0.88 & 0.87 & 0.23 & 0.60 & 0.82 \\
                           & 5 & 0.84 & 0.94 & 0.14 & 0.46 & 0.71 \\
    \midrule
    Structured dropout     & 1 & 0.93 & 0.61 & 0.41 & 0.78 & 0.91 \\
                           & 2 & 0.91 & 0.74 & 0.31 & 0.68 & 0.88 \\
                           & 3 & 0.89 & 0.88 & 0.19 & 0.55 & 0.74 \\
                           & 4 & 0.85 & 0.93 & 0.13 & 0.38 & 0.59 \\
                           & 5 & 0.81 & 0.98 & 0.04 & 0.14 & 0.22 \\
    \bottomrule
  \end{tabular}
\end{table}

\begin{table}[htbp]
  \centering
  \small
  \caption{Post-hoc CBM (multimodal mmL bank) ImageNet full stage-metric grid (ViT-B/16), giving Stage-1 cosine similarity, the standardized Stage-2 argmax-shift fraction, top-5 Jaccard, and Spearman $\rho$, and Stage-3 top-1 accuracy, for every corruption type and severity. Clean-reference accuracy $72.0\%$.}
  \label{tab:app:grid_pcbm}
  \begin{tabular}{llccccc}
    \toprule
    Corruption & S & cos\,$\uparrow$ & argmax\,$\downarrow$ & Jac\,$\uparrow$ & $\rho$\,$\uparrow$ & acc\,$\uparrow$ \\
    \midrule
    Motion blur            & 1 & 0.89 & 0.55 & 0.41 & 0.80 & 0.61 \\
                           & 2 & 0.84 & 0.65 & 0.32 & 0.71 & 0.53 \\
                           & 3 & 0.78 & 0.78 & 0.21 & 0.57 & 0.39 \\
                           & 4 & 0.72 & 0.88 & 0.12 & 0.43 & 0.25 \\
                           & 5 & 0.68 & 0.92 & 0.09 & 0.36 & 0.18 \\
    \midrule
    Gaussian noise         & 1 & 0.86 & 0.58 & 0.39 & 0.78 & 0.59 \\
                           & 2 & 0.82 & 0.66 & 0.31 & 0.71 & 0.50 \\
                           & 3 & 0.75 & 0.78 & 0.21 & 0.56 & 0.36 \\
                           & 4 & 0.66 & 0.89 & 0.11 & 0.37 & 0.21 \\
                           & 5 & 0.59 & 0.96 & 0.04 & 0.21 & 0.08 \\
    \midrule
    JPEG compression       & 1 & 0.88 & 0.55 & 0.42 & 0.81 & 0.62 \\
                           & 2 & 0.85 & 0.61 & 0.37 & 0.76 & 0.58 \\
                           & 3 & 0.84 & 0.63 & 0.34 & 0.74 & 0.54 \\
                           & 4 & 0.80 & 0.72 & 0.26 & 0.66 & 0.44 \\
                           & 5 & 0.74 & 0.81 & 0.18 & 0.54 & 0.31 \\
    \midrule
    Fog                    & 1 & 0.87 & 0.57 & 0.39 & 0.78 & 0.60 \\
                           & 2 & 0.85 & 0.61 & 0.36 & 0.74 & 0.57 \\
                           & 3 & 0.82 & 0.68 & 0.30 & 0.68 & 0.51 \\
                           & 4 & 0.80 & 0.73 & 0.25 & 0.62 & 0.46 \\
                           & 5 & 0.75 & 0.83 & 0.16 & 0.51 & 0.33 \\
    \midrule
    Random pixel dropout   & 1 & 0.87 & 0.59 & 0.38 & 0.76 & 0.60 \\
                           & 2 & 0.78 & 0.72 & 0.26 & 0.62 & 0.48 \\
                           & 3 & 0.73 & 0.78 & 0.20 & 0.54 & 0.40 \\
                           & 4 & 0.70 & 0.81 & 0.17 & 0.49 & 0.34 \\
                           & 5 & 0.67 & 0.85 & 0.14 & 0.43 & 0.28 \\
    \midrule
    Salt \& pepper         & 1 & 0.87 & 0.60 & 0.37 & 0.76 & 0.61 \\
                           & 2 & 0.80 & 0.72 & 0.26 & 0.65 & 0.50 \\
                           & 3 & 0.74 & 0.79 & 0.20 & 0.56 & 0.41 \\
                           & 4 & 0.70 & 0.83 & 0.15 & 0.49 & 0.33 \\
                           & 5 & 0.65 & 0.88 & 0.11 & 0.41 & 0.24 \\
    \midrule
    Sparse occlusion       & 1 & 0.89 & 0.51 & 0.45 & 0.79 & 0.67 \\
                           & 2 & 0.86 & 0.58 & 0.37 & 0.72 & 0.62 \\
                           & 3 & 0.82 & 0.65 & 0.31 & 0.66 & 0.57 \\
                           & 4 & 0.78 & 0.71 & 0.26 & 0.59 & 0.52 \\
                           & 5 & 0.75 & 0.76 & 0.22 & 0.53 & 0.44 \\
    \midrule
    Patch masking          & 1 & 0.91 & 0.47 & 0.48 & 0.83 & 0.68 \\
                           & 2 & 0.89 & 0.52 & 0.43 & 0.79 & 0.66 \\
                           & 3 & 0.86 & 0.57 & 0.39 & 0.75 & 0.64 \\
                           & 4 & 0.84 & 0.63 & 0.34 & 0.71 & 0.60 \\
                           & 5 & 0.81 & 0.70 & 0.27 & 0.64 & 0.52 \\
    \midrule
    Row/column dropout     & 1 & 0.91 & 0.51 & 0.45 & 0.82 & 0.67 \\
                           & 2 & 0.88 & 0.60 & 0.37 & 0.75 & 0.61 \\
                           & 3 & 0.85 & 0.67 & 0.30 & 0.69 & 0.54 \\
                           & 4 & 0.81 & 0.73 & 0.25 & 0.62 & 0.47 \\
                           & 5 & 0.77 & 0.77 & 0.21 & 0.55 & 0.40 \\
    \midrule
    Structured dropout     & 1 & 0.91 & 0.49 & 0.46 & 0.83 & 0.66 \\
                           & 2 & 0.88 & 0.55 & 0.41 & 0.79 & 0.63 \\
                           & 3 & 0.85 & 0.60 & 0.37 & 0.74 & 0.59 \\
                           & 4 & 0.81 & 0.69 & 0.29 & 0.66 & 0.52 \\
                           & 5 & 0.77 & 0.75 & 0.24 & 0.60 & 0.45 \\
    \bottomrule
  \end{tabular}
\end{table}

\begin{table}[htbp]
  \centering
  \small
  \caption{Post-hoc CBM (CAV bank from the $K{=}112$ labelled CUB attributes) CUB full stage-metric grid (ViT-B/16), giving Stage-1 cosine similarity, the standardized Stage-2 argmax-shift fraction, top-5 Jaccard, and Spearman $\rho$, and Stage-3 top-1 accuracy, for every corruption type and severity. Clean-reference accuracy $77.67\%$.}
  \label{tab:app:grid_pcbm_cub}
  \begin{tabular}{llccccc}
    \toprule
    Corruption & S & cos\,$\uparrow$ & argmax\,$\downarrow$ & Jac\,$\uparrow$ & $\rho$\,$\uparrow$ & acc\,$\uparrow$ \\
    \midrule
    Clean & 0 & 1 & 0 & 1 & 1 & 0.78 \\
    \midrule
    Motion blur            & 1 & 0.93 & 0.62 & 0.42 & 0.85 & 0.69 \\
                           & 2 & 0.88 & 0.72 & 0.33 & 0.77 & 0.59 \\
                           & 3 & 0.81 & 0.82 & 0.23 & 0.63 & 0.40 \\
                           & 4 & 0.73 & 0.91 & 0.16 & 0.48 & 0.23 \\
                           & 5 & 0.69 & 0.94 & 0.12 & 0.38 & 0.16 \\
    \midrule
    Gaussian noise         & 1 & 0.89 & 0.69 & 0.33 & 0.79 & 0.64 \\
                           & 2 & 0.85 & 0.78 & 0.26 & 0.70 & 0.53 \\
                           & 3 & 0.78 & 0.88 & 0.18 & 0.54 & 0.35 \\
                           & 4 & 0.69 & 0.95 & 0.11 & 0.34 & 0.16 \\
                           & 5 & 0.60 & 0.98 & 0.06 & 0.16 & 0.06 \\
    \midrule
    JPEG compression       & 1 & 0.86 & 0.69 & 0.33 & 0.78 & 0.62 \\
                           & 2 & 0.85 & 0.74 & 0.30 & 0.73 & 0.56 \\
                           & 3 & 0.84 & 0.76 & 0.28 & 0.70 & 0.51 \\
                           & 4 & 0.80 & 0.83 & 0.22 & 0.59 & 0.38 \\
                           & 5 & 0.75 & 0.89 & 0.16 & 0.45 & 0.23 \\
    \midrule
    Fog                    & 1 & 0.91 & 0.67 & 0.37 & 0.82 & 0.65 \\
                           & 2 & 0.89 & 0.73 & 0.33 & 0.78 & 0.60 \\
                           & 3 & 0.87 & 0.78 & 0.29 & 0.73 & 0.55 \\
                           & 4 & 0.86 & 0.79 & 0.27 & 0.71 & 0.53 \\
                           & 5 & 0.81 & 0.84 & 0.21 & 0.62 & 0.42 \\
    \midrule
    Random pixel dropout   & 1 & 0.89 & 0.64 & 0.37 & 0.81 & 0.67 \\
                           & 2 & 0.82 & 0.75 & 0.28 & 0.70 & 0.54 \\
                           & 3 & 0.76 & 0.82 & 0.22 & 0.62 & 0.44 \\
                           & 4 & 0.73 & 0.88 & 0.18 & 0.54 & 0.34 \\
                           & 5 & 0.69 & 0.92 & 0.14 & 0.45 & 0.25 \\
    \midrule
    Salt \& pepper         & 1 & 0.89 & 0.65 & 0.37 & 0.81 & 0.66 \\
                           & 2 & 0.83 & 0.74 & 0.29 & 0.71 & 0.55 \\
                           & 3 & 0.78 & 0.80 & 0.23 & 0.64 & 0.45 \\
                           & 4 & 0.74 & 0.85 & 0.19 & 0.56 & 0.36 \\
                           & 5 & 0.70 & 0.90 & 0.15 & 0.45 & 0.24 \\
    \midrule
    Sparse occlusion       & 1 & 0.86 & 0.67 & 0.36 & 0.81 & 0.69 \\
                           & 2 & 0.81 & 0.73 & 0.31 & 0.76 & 0.65 \\
                           & 3 & 0.78 & 0.78 & 0.27 & 0.73 & 0.61 \\
                           & 4 & 0.75 & 0.83 & 0.23 & 0.68 & 0.53 \\
                           & 5 & 0.73 & 0.87 & 0.19 & 0.61 & 0.42 \\
    \midrule
    Patch masking          & 1 & 0.89 & 0.61 & 0.41 & 0.84 & 0.71 \\
                           & 2 & 0.85 & 0.68 & 0.35 & 0.80 & 0.66 \\
                           & 3 & 0.82 & 0.74 & 0.30 & 0.75 & 0.59 \\
                           & 4 & 0.79 & 0.80 & 0.25 & 0.69 & 0.50 \\
                           & 5 & 0.75 & 0.87 & 0.19 & 0.56 & 0.35 \\
    \midrule
    Row/column dropout     & 1 & 0.90 & 0.61 & 0.43 & 0.86 & 0.73 \\
                           & 2 & 0.87 & 0.67 & 0.36 & 0.81 & 0.67 \\
                           & 3 & 0.84 & 0.75 & 0.30 & 0.75 & 0.59 \\
                           & 4 & 0.80 & 0.83 & 0.24 & 0.67 & 0.49 \\
                           & 5 & 0.77 & 0.88 & 0.19 & 0.61 & 0.40 \\
    \midrule
    Structured dropout     & 1 & 0.90 & 0.62 & 0.40 & 0.84 & 0.69 \\
                           & 2 & 0.88 & 0.68 & 0.35 & 0.80 & 0.65 \\
                           & 3 & 0.85 & 0.73 & 0.31 & 0.76 & 0.60 \\
                           & 4 & 0.81 & 0.81 & 0.25 & 0.68 & 0.49 \\
                           & 5 & 0.78 & 0.84 & 0.21 & 0.62 & 0.43 \\
    \bottomrule
  \end{tabular}
\end{table}

\begin{table}[htbp]
  \centering
  \small
  \caption{Post-hoc CBM (Broden CAV bank, $K{=}166$) CIFAR-10 full stage-metric grid (ViT-B/16), giving Stage-1 cosine similarity, the standardized Stage-2 argmax-shift fraction, top-5 Jaccard, and Spearman $\rho$, and Stage-3 top-1 accuracy, for every corruption type and severity. Clean-reference accuracy $94.8\%$. The CAV bank replaces the ImageNet-specific mmL vocabulary of ~\Cref{tab:app:grid_pcbm}, so cells are comparable within this table, not against ImageNet.}
  \label{tab:app:grid_pcbm_cifar10}
  \begin{tabular}{llccccc}
    \toprule
    Corruption & S & cos\,$\uparrow$ & argmax\,$\downarrow$ & Jac\,$\uparrow$ & $\rho$\,$\uparrow$ & acc\,$\uparrow$ \\
    \midrule
    Motion blur            & 1 & 0.97 & 0.52 & 0.45 & 0.86 & 0.92 \\
                           & 2 & 0.94 & 0.68 & 0.31 & 0.75 & 0.87 \\
                           & 3 & 0.92 & 0.78 & 0.22 & 0.64 & 0.80 \\
                           & 4 & 0.92 & 0.78 & 0.22 & 0.64 & 0.80 \\
                           & 5 & 0.89 & 0.85 & 0.16 & 0.55 & 0.73 \\
    \midrule
    Gaussian noise         & 1 & 0.88 & 0.80 & 0.21 & 0.58 & 0.84 \\
                           & 2 & 0.86 & 0.87 & 0.14 & 0.45 & 0.71 \\
                           & 3 & 0.84 & 0.92 & 0.09 & 0.34 & 0.57 \\
                           & 4 & 0.83 & 0.93 & 0.07 & 0.29 & 0.50 \\
                           & 5 & 0.82 & 0.95 & 0.06 & 0.25 & 0.44 \\
    \midrule
    JPEG compression       & 1 & 0.92 & 0.74 & 0.25 & 0.67 & 0.85 \\
                           & 2 & 0.89 & 0.83 & 0.18 & 0.57 & 0.78 \\
                           & 3 & 0.89 & 0.85 & 0.16 & 0.53 & 0.76 \\
                           & 4 & 0.88 & 0.87 & 0.14 & 0.48 & 0.73 \\
                           & 5 & 0.86 & 0.91 & 0.11 & 0.42 & 0.67 \\
    \midrule
    Fog                    & 1 & 0.99 & 0.31 & 0.64 & 0.95 & 0.95 \\
                           & 2 & 0.97 & 0.50 & 0.46 & 0.87 & 0.93 \\
                           & 3 & 0.95 & 0.62 & 0.36 & 0.79 & 0.91 \\
                           & 4 & 0.93 & 0.71 & 0.28 & 0.70 & 0.87 \\
                           & 5 & 0.89 & 0.84 & 0.16 & 0.52 & 0.74 \\
    \midrule
    Random pixel dropout   & 1 & 0.94 & 0.63 & 0.36 & 0.78 & 0.93 \\
                           & 2 & 0.91 & 0.77 & 0.24 & 0.64 & 0.88 \\
                           & 3 & 0.87 & 0.85 & 0.16 & 0.54 & 0.77 \\
                           & 4 & 0.85 & 0.91 & 0.11 & 0.43 & 0.60 \\
                           & 5 & 0.83 & 0.96 & 0.06 & 0.31 & 0.42 \\
    \midrule
    Salt \& pepper         & 1 & 0.94 & 0.63 & 0.36 & 0.78 & 0.93 \\
                           & 2 & 0.91 & 0.78 & 0.24 & 0.66 & 0.89 \\
                           & 3 & 0.88 & 0.85 & 0.16 & 0.54 & 0.79 \\
                           & 4 & 0.85 & 0.91 & 0.10 & 0.42 & 0.61 \\
                           & 5 & 0.82 & 0.97 & 0.06 & 0.28 & 0.41 \\
    \midrule
    Sparse occlusion       & 1 & 0.94 & 0.64 & 0.35 & 0.77 & 0.93 \\
                           & 2 & 0.92 & 0.75 & 0.25 & 0.68 & 0.90 \\
                           & 3 & 0.89 & 0.82 & 0.18 & 0.58 & 0.84 \\
                           & 4 & 0.86 & 0.88 & 0.13 & 0.49 & 0.71 \\
                           & 5 & 0.84 & 0.93 & 0.08 & 0.38 & 0.55 \\
    \midrule
    Patch masking          & 1 & 0.96 & 0.59 & 0.39 & 0.81 & 0.93 \\
                           & 2 & 0.94 & 0.67 & 0.33 & 0.75 & 0.91 \\
                           & 3 & 0.92 & 0.74 & 0.26 & 0.67 & 0.88 \\
                           & 4 & 0.90 & 0.84 & 0.18 & 0.57 & 0.78 \\
                           & 5 & 0.86 & 0.92 & 0.11 & 0.43 & 0.64 \\
    \midrule
    Row/column dropout     & 1 & 0.96 & 0.59 & 0.39 & 0.82 & 0.92 \\
                           & 2 & 0.93 & 0.71 & 0.29 & 0.73 & 0.90 \\
                           & 3 & 0.91 & 0.79 & 0.22 & 0.65 & 0.87 \\
                           & 4 & 0.88 & 0.87 & 0.14 & 0.52 & 0.78 \\
                           & 5 & 0.84 & 0.94 & 0.08 & 0.38 & 0.64 \\
    \midrule
    Structured dropout     & 1 & 0.93 & 0.71 & 0.29 & 0.72 & 0.92 \\
                           & 2 & 0.91 & 0.79 & 0.22 & 0.63 & 0.87 \\
                           & 3 & 0.89 & 0.88 & 0.14 & 0.51 & 0.74 \\
                           & 4 & 0.85 & 0.95 & 0.07 & 0.35 & 0.53 \\
                           & 5 & 0.81 & 0.99 & 0.03 & 0.13 & 0.25 \\
    \bottomrule
  \end{tabular}
\end{table}

\paragraph{Per-corruption type prediction instability.}
The shift rate, the fraction of adjacent-severity steps at which an image's top-1 label changes as severity climbs from~1 to~5, matches the accuracy ordering of ~\Cref{sec:propagation}, where global-statistic corruptions flip labels most, local occluders least, and every corruption type flips the label more often on ImageNet than on CIFAR-10.

\begin{figure}[htbp]
  \centering
  \includegraphics[width=0.8\textwidth]{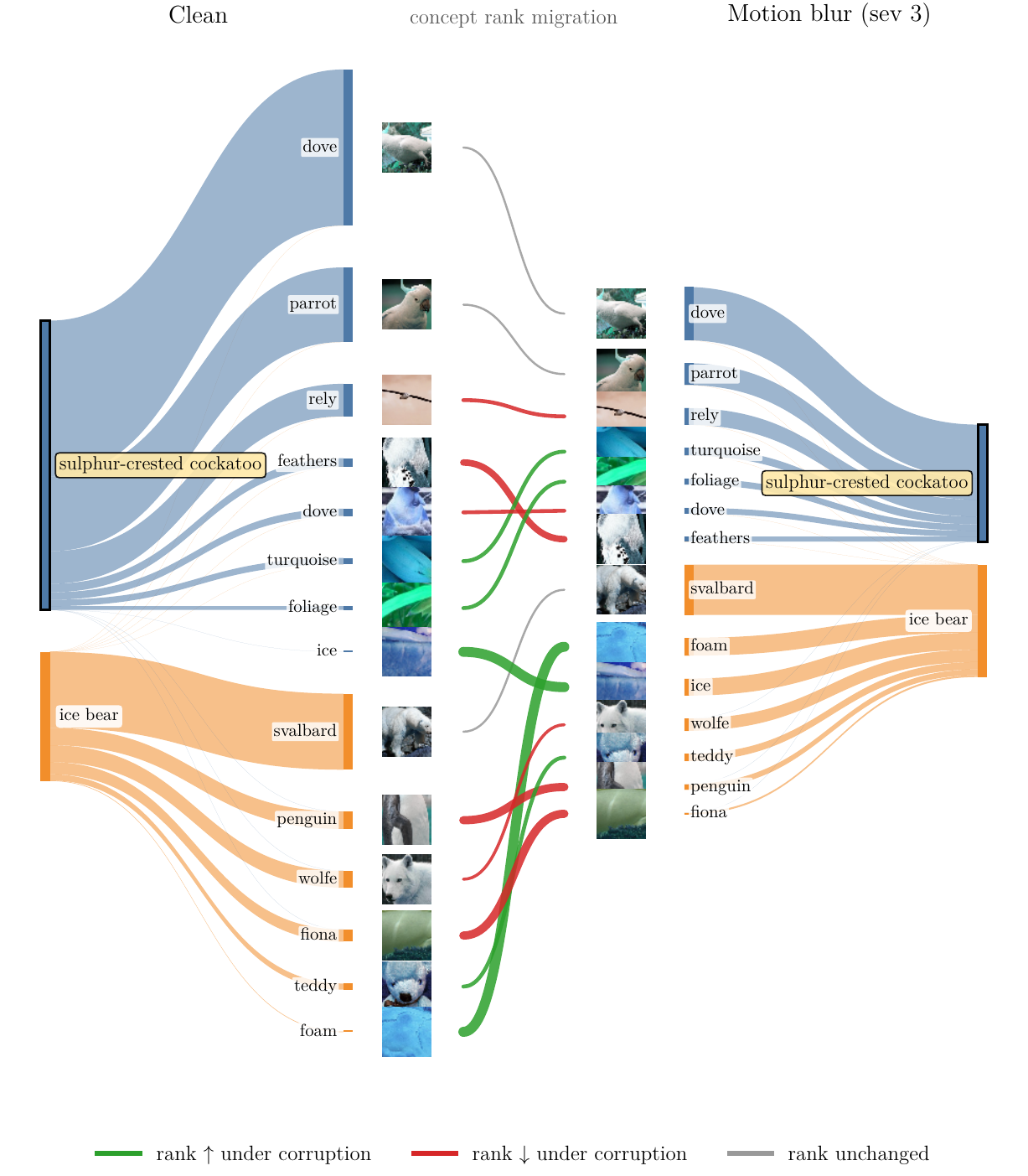}
  \caption{Concept-to-prediction flow for the cockatoo image of ~\Cref{fig:flow}, clean (left) versus motion blur at severity~$3$ (right), same encoding. The ranking reshuffles across the bridge, yet the correct class still wins, a reordering without a change of prediction.}
  \label{fig:app:motionblur}
\end{figure}

\paragraph{Reordering without a change of prediction.}
The pixel-dropout example of ~\Cref{fig:flow} shows a reordering that flips the label. ~\Cref{fig:app:motionblur} is its complement. Under motion blur the same image's concept ranking reshuffles just as visibly, yet the reshuffled contributions still sum through the linear head to a winning margin for the correct class. Stage-2 reordering is therefore necessary but not sufficient for a misprediction, and whether it crosses the decision boundary depends on how the head aggregates the shuffled concepts.

\section{The Per-Concept Shift \texorpdfstring{$d_k$}{dk}}
\label{app:dk}

\subsection{Definition and Per-Model Transfer}
\label{app:dk:def}
\Cref{eq:dk} defines $d_k$ for DCBM on the pre-softmax concept score $\bar a_k(x)$, averaged over the ten corruption types $F$ and five severities $S$ against the clean score $\bar a^{\,\mathrm{cl}}_k(x)$. The estimator transfers unchanged to any CBM whose concept layer produces a concept activation score, which is what makes the genericity replication possible, since only the definition of the score $\bar a_k(x)$ changes between models,
\begin{align}
  \text{DCBM:}   \quad \bar a_k(x) &= \frac{\langle z(x), c_k\rangle}{\lVert c_k\rVert_2^{2}}, \nonumber\\
  \text{LF-CBM:} \quad \bar a_k(x) &= \langle w_k, z(x)\rangle, \nonumber\\
  \text{PCBM:}   \quad \bar a_k(x) &= \frac{\langle z(x), c_k\rangle}{\lVert c_k\rVert_2^{2}},
  \label{eq:dk:families}
\end{align}
with $z(x)$ the frozen CLIP feature, $c_k$ the concept medoids (DCBM) or CLIP-text concept vector (PCBM), and $w_k$ the learned text-projection weight (LF-CBM).
We define $d_k$ on scores rather than post-softmax activations because the $K$-way normalizer couples all coordinates, so a per-concept statistic on activations would largely reflect the normalizer. All downstream accuracies nonetheless use the true forward path (softmax, then scaler, then head). ~\Cref{fig:dkoverview} shows DCBM's distribution, the band structure shared across the ten corruption types, and the monotone severity response.

\paragraph{Relative shift for PCBM on CUB.}
$d_k$ is measured in the unit of the concept score, so it compares two concepts only where the concept layer gives them a common scale, which holds for the crop bank of DCBM, the scale-free projection of LF-CBM and the $L_2$-normalized text vectors of PCBM on ImageNet.
On CUB the concept vectors of PCBM are the normals of linear classifiers fit on the attribute labels, whose norm the fit leaves free, and the score of~\Cref{eq:dk:families} divides by the squared norm, so the scale of a concept follows the norm of its vector.
For PCBM on CUB, we therefore rank by the relative shift
\begin{equation}
  \tilde d_k = d_k / \sigma_k ,
  \label{eq:dk:relative}
\end{equation}
with $\sigma_k$ the standard deviation of the clean score of concept $k$ over the training images, which is the shift of the standardized activation $\tilde a_k$ the head reads; every other model is ranked by $d_k$.
Rows and curves ranked by $\tilde d_k$ are marked $^{\ddagger}$.

\subsection{The Distribution of \texorpdfstring{$d_k$}{dk} Across Models}

\begin{figure}[htbp]
  \centering
  \includegraphics[width=\textwidth]{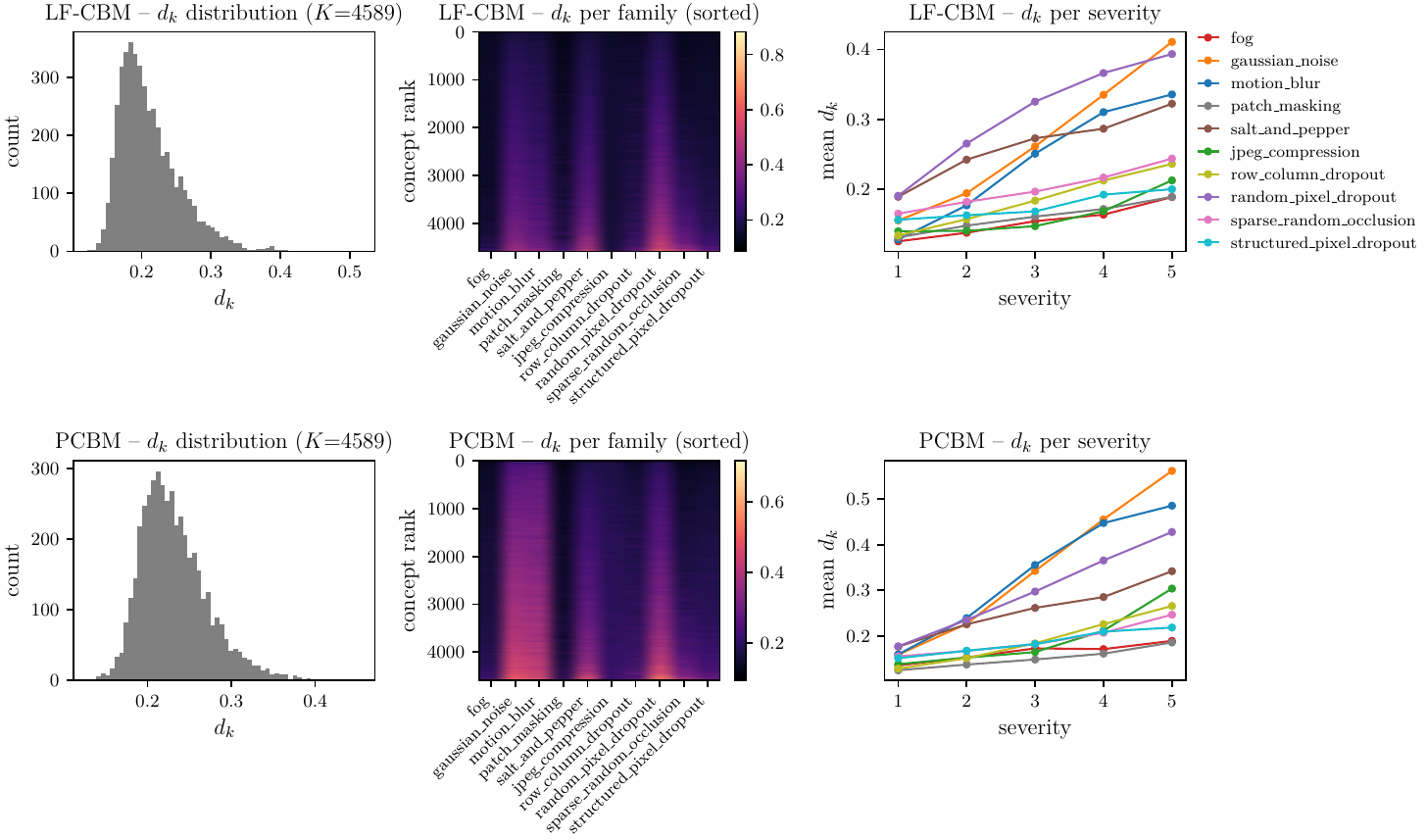}
  \caption{Per-concept shift $d_k$ distributions for Label-free CBM and the multimodal Post-hoc CBM, each unimodal with a long right tail, matching the DCBM structure of ~\Cref{fig:dkoverview}.}
  \label{fig:app:genericitydk}
\end{figure}

~\Cref{fig:dkoverview} in ~\Cref{sec:robust} shows the DCBM distribution. The distribution shape reported for DCBM in ~\Cref{sec:robust} is not specific to image-crop banks. Computing $d_k$ from the per-model scores of~\Cref{eq:dk:families} on Label-free CBM and the multimodal Post-hoc CBM gives the same unimodal, long-right-tailed form on each $4589$-concept bank, so a robust subset is well defined for each model (\Cref{fig:app:genericitydk}).

\subsection{Three Estimates of \texorpdfstring{$d_k$}{dk}}
\label{app:dk:estimates}
\label{app:dk:disjoint}
\label{app:dk:loco}
\Cref{eq:dk} is computed on the train split, i.e., $50$ per class on ImageNet and the entire training split on CUB.
We ablate two additional scenarios for computing $d_k$:
the \emph{test-pool} $d_k$ is computed on test images, assuming no access to the model's training set; writing $\mathcal{D}_{\mathrm{test}}$ for that pool, it keeps every other term of~\Cref{eq:dk},
\begin{equation}
  d_k^{\mathrm{test}} = \frac{1}{|F|\,|S|}\sum_{f\in F}\sum_{s\in S}\mathbb{E}_{x\in\mathcal{D}_{\mathrm{test}}}\bigl|\,\bar a^{\,f,s}_k(x)-\bar a^{\,\mathrm{clean}}_k(x)\,\bigr|.
  \label{eq:dk:disjoint}
\end{equation}
The \emph{leave-one-corruption-out} $d_k$ (LOCO) ablates the selection based on a subset (9/10) of corruptions. This computes $d_k$ based on the training set, as in the main paper.
For each corruption type $c\in F$, it is computed separately 
\begin{equation}
  d_k^{(-c)} = \frac{1}{(|F|-1)\,|S|}\sum_{f\in F\setminus\{c\}}\sum_{s\in S}\mathbb{E}_{x\in\mathcal{D}_{\mathrm{train}}}\bigl|\,\bar a^{\,f,s}_k(x)-\bar a^{\,\mathrm{clean}}_k(x)\,\bigr|,
  \label{eq:dk:loco}
\end{equation}
the $K$ lowest concepts are kept, the head is retrained, and the pruned head is scored on $c$, which the ranking never saw; the reported value is the mean over the ten choices of $c$.
We call the three estimates the train, the test-pool, and the LOCO $d_k$.
The three rankings agree closely, with a Spearman correlation of $0.97$ or more between the train and the test-pool ranking and $0.99$ or more between the LOCO and the train ranking on ImageNet, and $0.998$ or more on CUB, while a ranking from one corruption type alone correlates with the averaged ranking at only $0.65$ to $0.71$.
~\Cref{tab:app:estimates} scores the three subsets on the held-out corruption.
The estimate moves the accuracy of the pruned head by at most $1.5$ points, and the LOCO subset keeps the margin of the train subset over the deployed head for every model on both datasets.
The three pruning strategies differ by $0.002$-$0.015$ and have the same relation to the deployed model, i.e., higher corrupt accuracy than the deployed head for all three models on both datasets.
Thus, $d_k$ pruning generalizes across different selection strategies.

\begin{table}[htbp]
  \centering
  \caption{Three estimates of $d_k$ scored on the held-out corruption.
  \emph{Train} $d_k$ is the estimate used throughout the paper, computed on paired training images (50 per class on ImageNet, the full training split on CUB); \emph{test-pool} $d_k$ is computed on the test images instead, assuming no access to the training set (\Cref{eq:dk:disjoint}); \emph{LOCO} $d_k$ is computed on the training images from nine of the ten corruption types~(\Cref{eq:dk:loco}).
  Each column is the head retrained on the $K$ concepts one estimate ranks most stable, scored on the corruption type the LOCO estimate left out and averaged over the ten choices of that type; the deployed head is scored the same way.
  $K$ is $K^\star$ of ~\Cref{tab:plateau}; the train column corresponds to $\Delta\,d_k$ there.
  Severity-3 top-1 as fractions, head retrained with the model's recipe, three seeds.
  $^{\ddagger}$all three estimates of the relative shift $\tilde d_k$.}
  \label{tab:app:estimates}
  \begin{tabular}{llcccc}
    \toprule
    & Model ($K/K_{\text{full}}$) & Deployed & Train $d_k$ (main) & Test-pool $d_k$ & LOCO $d_k$ \\
    \midrule
    ImageNet & DCBM (1200/2048)           & $0.451$ & $0.464$ & $0.461$ & $0.463$ \\
             & LF-CBM (3600/4589)         & $0.532$ & $0.543$ & $0.543$ & $0.541$ \\
             & PCBM (3600/4589)           & $0.497$ & $0.504$ & $0.501$ & $0.501$ \\
    \midrule
    CUB      & DCBM (256/1024)            & $0.351$ & $0.446$ & $0.441$ & $0.441$ \\
             & LF-CBM (206/329)           & $0.397$ & $0.428$ & $0.430$ & $0.415$ \\
             & PCBM$^{\ddagger}$ (84/112) & $0.508$ & $0.526$ & $0.525$ & $0.518$ \\
    \bottomrule
  \end{tabular}
\end{table}

\subsection{Head Protocol for a Pruned Concept Set}
\label{app:protocol}
A pruned concept set needs a new head, because the deployed head carries one weight per concept of the full set and was fit to all of them together, and the head protocol decides what a difference between a pruned head and the deployed head measures.
A protocol whose head on the full set is weaker than the deployed one lets a pruned head win by a wide margin without beating the deployed model, and a stronger one lets it win without help from the pruning.
We therefore ask of every protocol that its head on the full set reproduce the deployed head on clean and on corrupted images, the reproduction gate, and read the pruning gain, the pruned head minus the deployed head, only under a protocol that passes it.
~\Cref{tab:app:protocols} compares six protocols.
\emph{Zero-out} keeps the deployed head and sets each pruned concept's standardized activation to its clean-train mean; nothing is trained, and accuracy collapses, since each deployed head is co-adapted to its full set (the LF-CBM and PCBM heads are about $1\%$ non-zero).
The \emph{$L_2$ refit} fits a fresh multinomial logistic head with an $L_2$ penalty on the $128{,}118$ held-out ImageNet training images, or the CUB training split, with one recipe for all three models, at the fixed $C{=}1$, and at a $C$ tuned on corrupted training images.
The \emph{own-penalty refit} keeps each model's own regularizer ($L_1$ for DCBM, elastic net for LF-CBM and PCBM) on the same ImageNet pool, at the deployed strength and at a tuned one.
The \emph{retrain} fits the head from scratch with the recipe of the model's deployed head (\Cref{tab:app:recipes}) on the training split; the CLIP backbone and the concept vectors stay frozen, pruned concepts are deleted from the concept layer, DCBM's softmax is recomputed over the kept concepts with the standardization re-fit, and no hyperparameter is re-tuned per $K$.
Only the retrain passes the gate for all three models on both datasets, so it is the protocol of the paper and the one a deployer can use, since its full bank head is the deployed head and a pruned head differs from it only in the pruning.
The fixed-$C=1$ refit sits $4.4$ to $10.6$ points below the deployed head on corrupted ImageNet images at the full set, so its pruning margins were read against a weakened reference; the direction of the selection result, the $d_k$ subset above a random subset of the same size on corrupted images, holds under every protocol on both datasets.

\begin{table}[htbp]
  \centering
  \caption{Head-retrain recipes, each the recipe of the model's deployed head.
  On ImageNet all three train on the same $1{,}153{,}049$ training images; on CUB on the $5{,}394$ training images, where DCBM uses batch $32$ with the head selected on the $600$-image validation split and LF-CBM and PCBM use $\lambda{=}2\cdot10^{-4}$ with $1000$ and $60$ epochs; seeds $\{0,1,2\}$.}
  \label{tab:app:recipes}
  \small
  \begin{tabular}{llll}
    \toprule
    & DCBM & LF-CBM & PCBM \\
    \midrule
    Initialization & fresh linear layer & zero & zero \\
    Optimizer      & Adam, lr $10^{-4}$ & GLM-SAGA, step $0.1$ & GLM-SAGA, step $0.1$ \\
    Objective      & CE $+\,10^{-4}\lVert W\rVert_1$ & elastic net, $\alpha{=}0.99$, $\lambda{=}10^{-4}$ & elastic net, $\alpha{=}0.99$, $\lambda{=}2\cdot10^{-4}$ \\
    Epochs, batch  & $200$, $512$ & $80$, $256$ & $60$, $256$ \\
    Head selection & best held-out loss & final path point & final path point \\
    \bottomrule
  \end{tabular}
\end{table}

\begin{table}[htbp]
  \centering
  \footnotesize
  \caption{Reproduction and ablation of six head protocols.
  We report the model's full bank head minus the deployed head, clean/corrupted top-1 in pp, closer to $0$ is better.
  Corrupted is the severity-3 mean over the ten corruption types; three seeds for the fitted protocols; tuned strengths are chosen on corrupted training images and kept across sizes.
  We compare different settings: training data available (retrain) and unavailable (refit).
  Only the retrain passes the gate for every model on both datasets.}
  \label{tab:app:protocols}
  \setlength{\tabcolsep}{4pt}
  \begin{tabular}{llllrrr}
    \toprule
    & & & & \multicolumn{3}{c}{full $-$ deployed, clean / corr.\ (pp)} \\
    \cmidrule(lr){5-7}
    Dataset & Protocol & Penalty & Fit on & DCBM & LF-CBM & PCBM \\
    \midrule
    ImageNet & zero-out & \multicolumn{2}{l}{deployed head, no fit} & $0.0$ / $0.0$ & $0.0$ / $0.0$ & $-0.1$ / $0.0$ \\
     & $L_2$ refit & $L_2$, $C{=}1$ & test & $+0.7$ / $-4.4$ & $-4.6$ / $-10.6$ & $-1.2$ / $-7.1$ \\
     & $L_2$ refit & $L_2$, tuned $C$ & test & $+3.7$ / $+0.8$ & $-1.2$ / $-2.6$ & $+2.0$ / $+1.0$ \\
     & own-penalty refit & own, deployed $\lambda$ & test & $-1.2$ / $-1.8$ & $-1.1$ / $-1.3$ & $0.0$ / $+0.5$ \\
     & own-penalty refit & own, tuned $\lambda$ & test & $+3.4$ / $+1.1$ & $-0.6$ / $-1.9$ & $+2.7$ / $+1.3$ \\
     & retrain & own, deployed $\lambda$ & train & $-0.1$ / $+0.1$ & $0.0$ / $0.0$ & $0.0$ / $0.0$ \\
    \midrule
    CUB & zero-out & \multicolumn{2}{l}{deployed head, no fit} & $0.0$ / $0.0$ & $0.0$ / $0.0$ & $0.0$ / $0.0$ \\
     & $L_2$ refit & $L_2$, $C{=}1$ & train & $-2.8$ / $-3.2$ & $-1.1$ / $-3.5$ & $-1.1$ / $-4.6$ \\
     & $L_2$ refit & $L_2$, tuned $C$ & train & $-1.5$ / $-1.0$ & $-2.0$ / $+2.5$ & $-2.0$ / $-0.4$ \\
     & own-penalty refit & own, deployed $\lambda$ & train & -- & -- & -- \\
     & own-penalty refit & own, tuned $\lambda$ & train & -- & -- & -- \\
     & retrain & own, deployed $\lambda$ & train & $-0.2$ / $+0.5$ & $0.0$ / $0.0$ & $0.0$ / $0.0$ \\
    \bottomrule
  \end{tabular}
\end{table}

\subsection{Concept Selection Across Models}
\label{app:genericity}
\begin{figure}[htbp]
  \centering
  \includegraphics[width=0.8\textwidth]{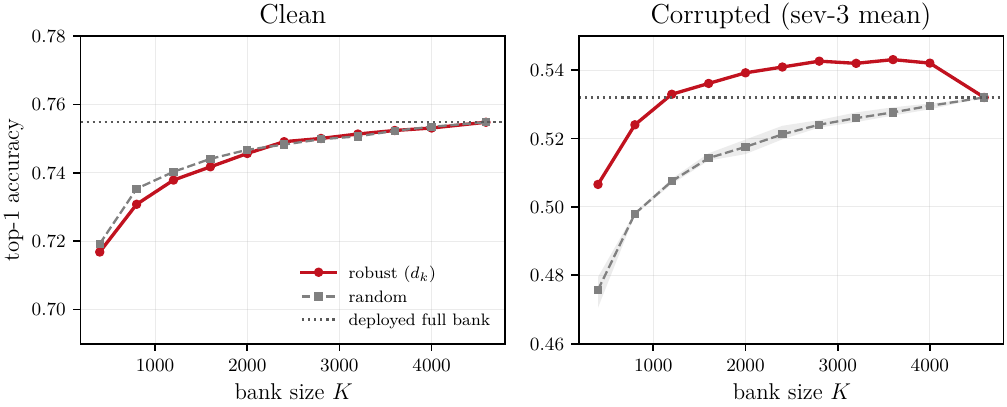}
  \caption{Capacity-matched $d_k$-versus-random selection on Label-free CBM (CLIP ViT-B/16, head retrained with the LF-CBM recipe, three seeds at $K\in\{400,800,3600\}$ and one seed elsewhere). The robust arm beats the random subset on corrupted accuracy across capacities (peak $+3.1$ points at $K{=}400$), and the margin decays toward zero as the bank grows. Clean accuracy stays within $0.5$ points of random.}
  \label{fig:app:lfcbm}
\end{figure}

\begin{figure}[htbp]
  \centering
  \includegraphics[width=0.8\textwidth]{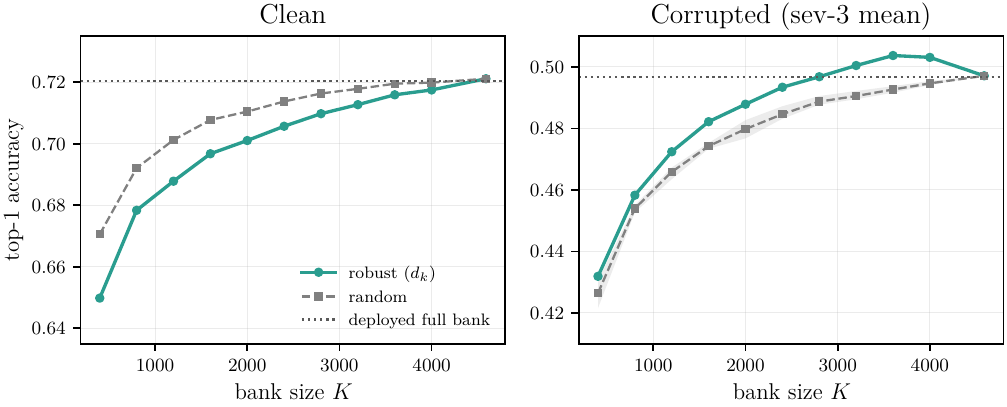}
  \caption{Capacity-matched $d_k$-versus-random selection on the multimodal Post-hoc CBM (CLIP ViT-B/16, head retrained with the PCBM recipe, three seeds at $K\in\{800,3600,4000\}$ and one seed elsewhere); the robust arm beats the random null on corrupted accuracy (peak $+1.1$ points at $K{=}3600$) and the margin decays toward zero as the bank grows; clean accuracy sits $0.2$ to $2.1$ points below random.}
  \label{fig:app:pcbm}
\end{figure}

The concept-selection result of~\Cref{sec:robust} rests on replicating the capacity-matched prune-versus-random margin on both text-bank models. 
~\Cref{fig:app:lfcbm,fig:app:pcbm} give the per-$K$ selection curves on ImageNet, each head retrained with the model's recipe (\Cref{app:protocol}). The robust $d_k$-pruned subsets beats the size-matched random subset on corrupted accuracy across for all choices of $K$.
The CUB curves for all three models are the bottom row of ~\Cref{fig:genericity}.

\subsection{DCBM Case Study: Capacity and Crop Geometry}
\label{app:casestudy}
\begin{figure}[htbp]
  \centering
  \includegraphics[width=0.8\textwidth]{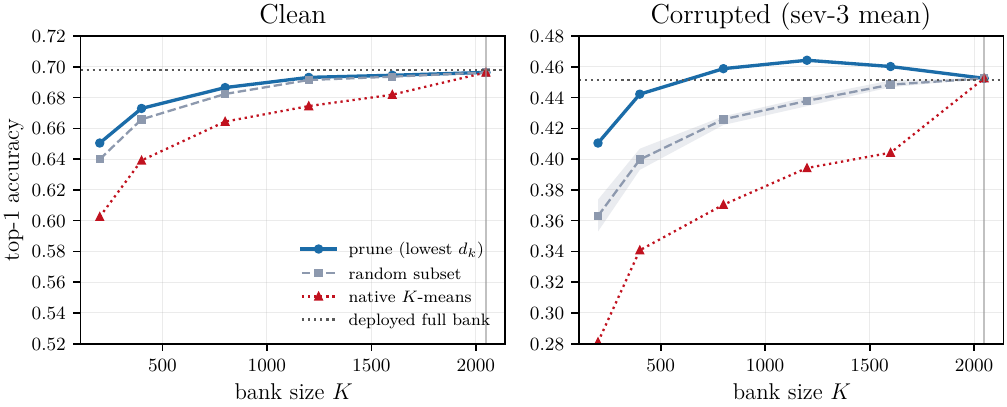}
  \caption{Capacity-matched selection on ImageNet DCBM, each head retrained with the deployed DCBM recipe.
  At each bank size $K$, pruning the $K{=}2048$ bank to its $K$ lowest-$d_k$ concepts beats a size-matched random subset by $+1.2$ to $+4.7$ points of corrupted accuracy and a natively built $K$-means bank by $+5.6$ to $+12.9$ points, peaking at $K{=}200$ ($+4.7$ over random, $+12.9$ over native).
  From $K{=}800$ to $1600$ the pruned bank exceeds the deployed full-bank head (dotted) on corrupted accuracy, by up to $+1.3$ points at $K{=}1200$.}
  \label{fig:capacity}
\end{figure}

DCBM's concept bank is built from image crops rather than a pre-filtered text vocabulary, which raises two DCBM-specific questions about what its $d_k$ margin measures.
First, for a crop bank we can also build a \emph{native} bank, a fresh $K$-means clustering of size $K$ from the full crop pool, and compare it with pruning the full $K{=}2048$ bank to the same size (\Cref{fig:capacity}).
Pruning wins by $+5.6$ to $+12.9$ points of corrupted accuracy ($+12.9$ at $K{=}200$), so building a large bank and pruning it beats building a small one directly, as the pruning literature finds for subnetworks of large trained models~\citep{frankle_lottery_2019}.

Second, we ask what the ranking captures in image space.
If fragile concepts are simply those built from small crops, $d_k$ might rank crop-acquisition quality rather than robustness.
Three tests (\Cref{app:geometry} below) agree it is both, with neither part negligible.
Correlationally, $d_k$ falls with the medoid crop's short side (Spearman $\rho=-0.56$, ~\Cref{fig:app:dkgeom}).
Causally, downsampling only a robust concept's crop and re-embedding lifts its $d_k$ by a median $+0.061$ (\Cref{tab:dkcausal}).
Yet the $d_k$ and resolution rankings keep largely different concepts (Jaccard $0.13$--$0.36$, ~\Cref{fig:app:overlap}), so resolution is no substitute.
At a bank of $200$ concepts the $+4.7$-point margin of $d_k$ over random splits into $+2.5$ points from crop resolution alone, free at build time, and $+2.2$ beyond geometry, with a resolution-residualized arm still beating random by $+3.1$ (\Cref{fig:app:pruninggeom}).
This is why DCBM's margin over random is the largest of the three models, as it carries both concept stability and crop geometry.

\subsection{What \texorpdfstring{$d_k$}{dk} Measures: Crop Geometry (DCBM)}
\label{app:geometry}
\begin{figure}[htbp]
  \centering
  \includegraphics[width=\textwidth]{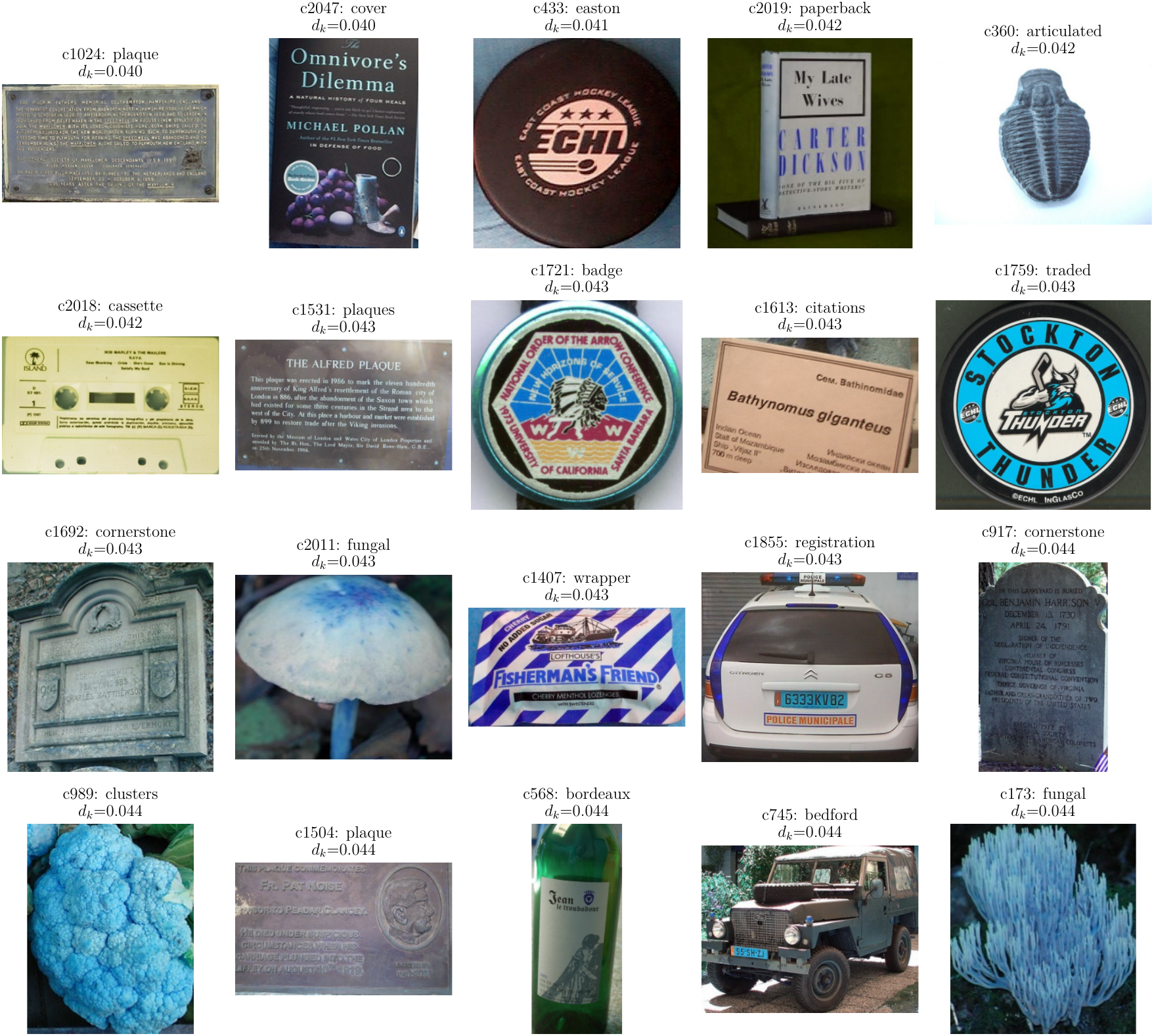}
  \caption{Medoid crops of the most robust (lowest-$d_k$) concepts. Robust concepts tend to be large, sharp, roughly square crops of whole objects.}
  \label{fig:app:robust}
\end{figure}

\begin{figure}[htbp]
  \centering
  \includegraphics[width=\textwidth]{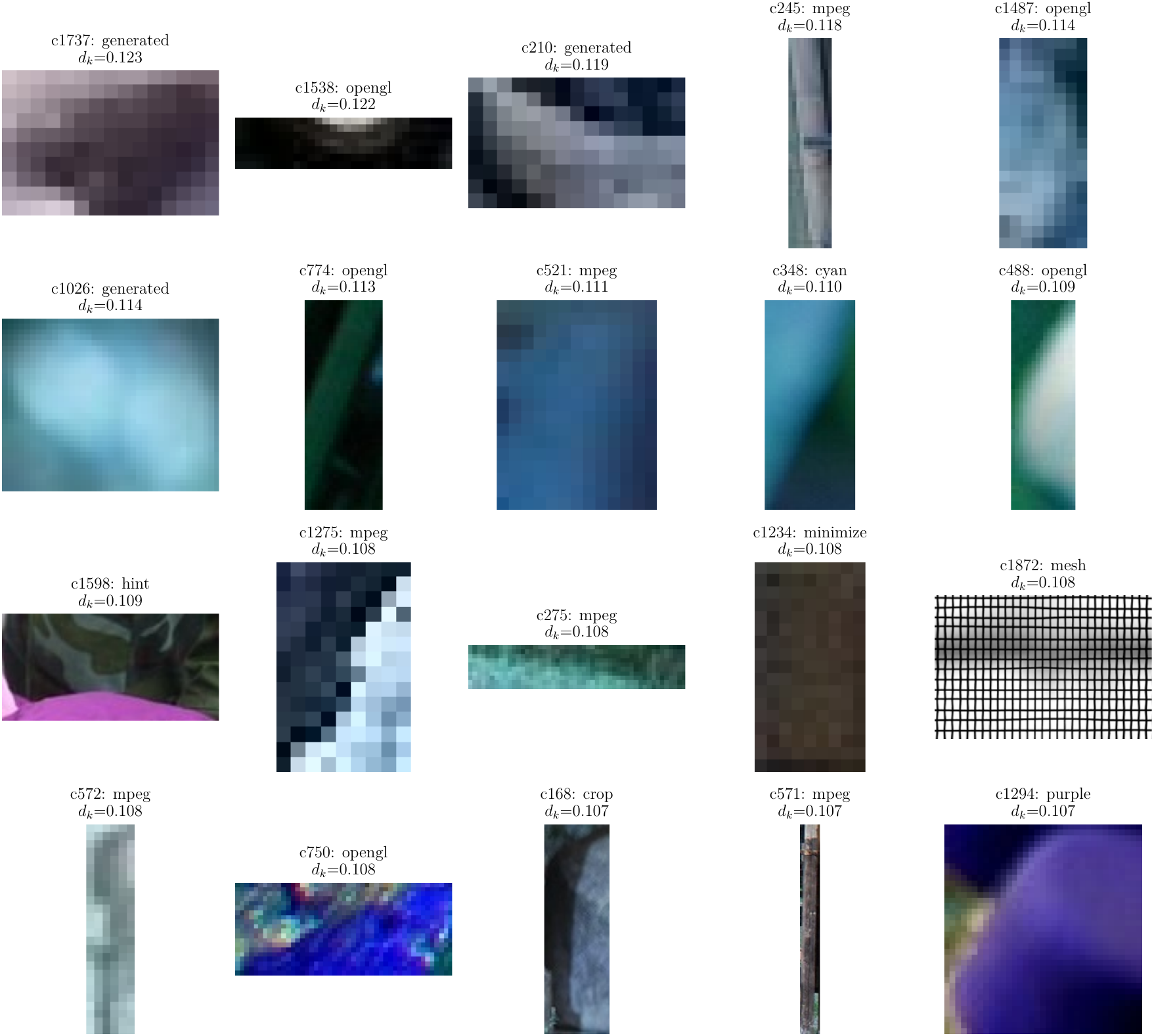}
  \caption{Medoid crops of the most fragile (highest-$d_k$) concepts, typically small, thin, low-resolution slivers. The geometry contrast with ~\Cref{fig:app:robust} is the confound examined in this subsection.}
  \label{fig:app:fragile}
\end{figure}

Ranking DCBM concepts by $d_k$ exposes a visible geometry contrast, the most robust concepts being large, sharp, roughly square crops of whole objects (\Cref{fig:app:robust}) and the most fragile small, thin, low-resolution slivers (\Cref{fig:app:fragile}). This raises the possibility that $d_k$ ranks crop-acquisition quality rather than robustness. We test this correlationally, causally, and downstream, on all $618{,}743$ SAM2 crops (width and height read from JPEG headers without decoding).

\begin{figure}[t]
  \centering
  \includegraphics[width=\textwidth]{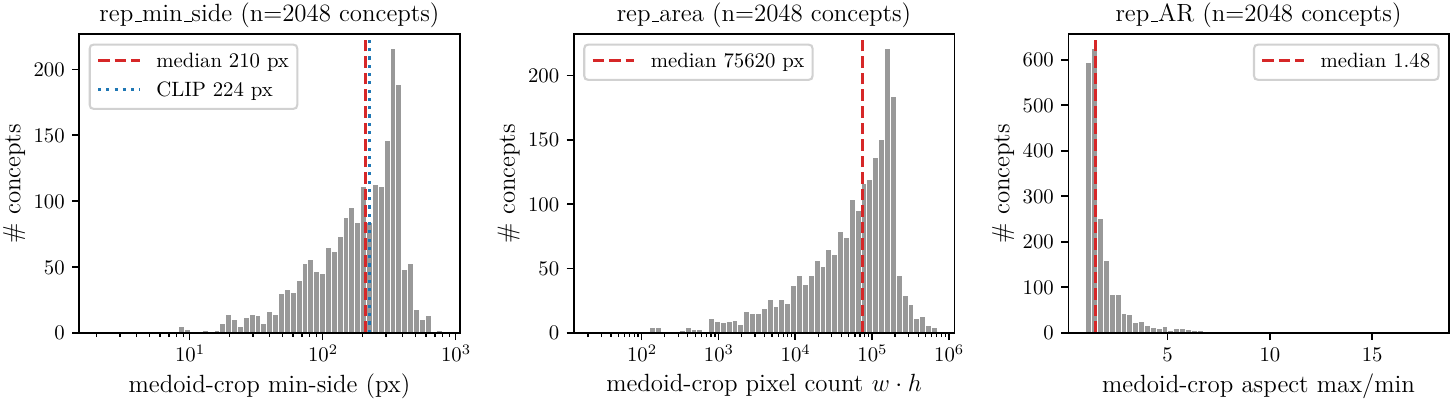}
  \caption{Distribution of medoid-crop geometry over the $2048$ DCBM concepts, read from JPEG headers of $618{,}743$ crops. The short side binds CLIP's $224$-pixel resize; its median is $210$ pixels.}
  \label{fig:cropdist}
\end{figure}

\begin{figure}[htbp]
  \centering
  \includegraphics[width=\textwidth]{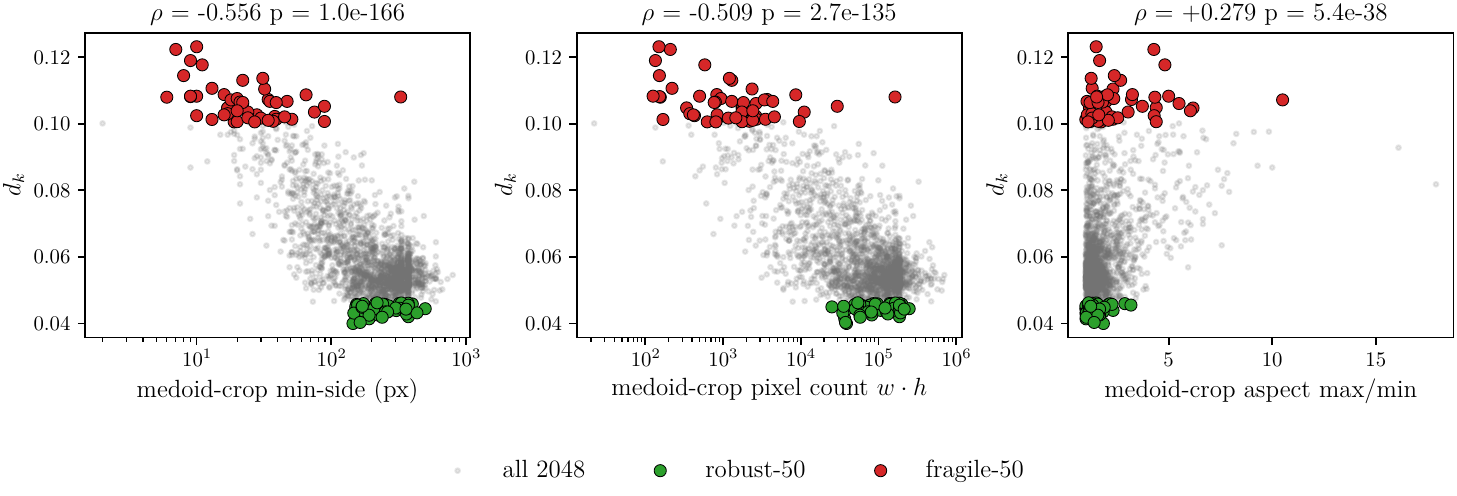}
  \caption{$d_k$ against each medoid-crop geometry measure. Resolution correlates with the shift ($\rho=-0.56$), the effect strong for high-frequency corruptions and weak for occlusion and dropout.}
  \label{fig:app:dkgeom}
\end{figure}

\begin{figure}[htbp]
  \centering
  \includegraphics[width=0.8\textwidth]{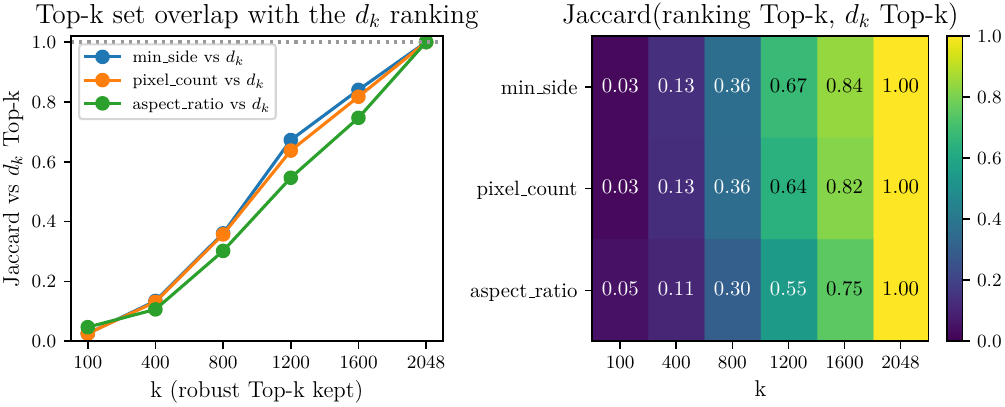}
  \caption{Top-$k$ Jaccard overlap between the robust-first $d_k$ ranking and each geometry ranking. At $k\in\{400,800\}$ the short-side ranking overlaps $d_k$ by only $0.13$--$0.36$ despite the $\rho=-0.56$ correlation.}
  \label{fig:app:overlap}
\end{figure}

\begin{figure}[htbp]
  \centering
  \includegraphics[width=0.8\textwidth]{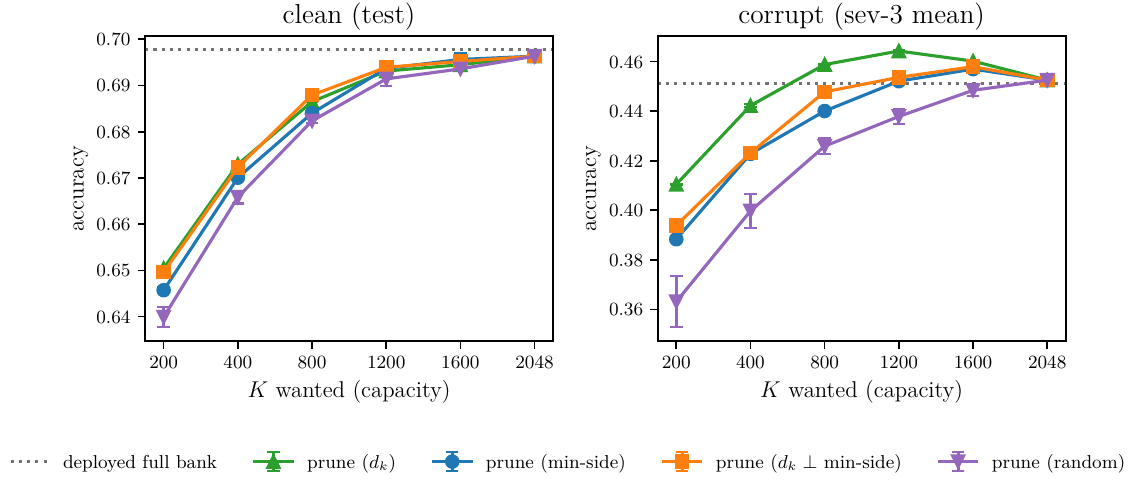}
  \caption{Corrupted-accuracy margin over a size-matched random subset (pp) for the $d_k$, geometry-only (largest medoid crop first), and resolution-residualized $d_k$ prune rules, capacity-matched across the $K$ spectrum, head retrained with the DCBM recipe.
  The geometry-only rule is corruption-free and available at build time; the residualized arm regresses resolution out of $d_k$.
  At $K{=}200$ the $d_k$ margin ($+4.7$) splits into $+2.5$ from crop resolution alone and $+2.2$ beyond geometry, with the residualized arm at $+3.1$; the $d_k$ margin collapses at the full $K{=}2048$ bank, where the arms coincide.}
  \label{fig:app:pruninggeom}
\end{figure}

The binding quantity is the medoid crop's short side, because CLIP resizes the short side to $224$ pixels. Its median is $210$ pixels, so about half the concepts ($1083/2048$) are built from crops CLIP upsamples, yielding low-frequency embeddings (\Cref{fig:cropdist}). Correlationally, $d_k$ falls with the short side at Spearman $\rho=-0.56$ (partial $\rho=-0.46$ controlling for medoid norm), strongly for the high-frequency noise and blur corruptions and weakly for occlusion and dropout (\Cref{fig:app:dkgeom}). Causally, downsampling only a robust concept's medoid crop to the fragile-set median short side ($16$ px) and re-embedding lifts its $d_k$ by a median $+0.061$ ($0.044\!\to\!0.106$), more than the $+0.045$ the same treatment adds to mid-$d_k$ concepts, confirming that lower resolution causally raises the shift, and a control that re-embeds the crop unchanged does not move (\Cref{tab:dkcausal}). Yet at the size $K\in\{400,800\}$ the $d_k$ and short-side rankings overlap only $0.13$--$0.36$ in Jaccard (\Cref{fig:app:overlap}), so resolution is not a substitute for $d_k$. 
The downstream decomposition settles it (\Cref{fig:app:pruninggeom}), read with the head retrained under the DCBM recipe. At $K{=}200$ the $+4.7$-point margin over random splits into $+2.5$ points from crop resolution alone, free at build time, and $+2.2$ points beyond geometry, with a resolution-residualized $d_k$ arm still beating random by $+3.1$ points.

\begin{table}[htbp]
  \centering
  \caption{Causal crop-resolution test: median change in $d_k$ after downsampling only the medoid crop to the fragile-set median short side ($16$ px) and re-embedding, with the shift averaged over severities $1$, $3$ and $5$ of the training pool. Low resolution raises the measured shift, more for robust than for mid-$d_k$ concepts; an unchanged re-embedding control does not move.}
  \label{tab:dkcausal}
  \begin{tabular}{lc}
    \toprule
    Treatment & Median $\Delta d_k$ \\
    \midrule
    Robust concepts, downsampled to $16$ px    & $+0.061$ ($0.044\!\to\!0.106$) \\
    Mid-$d_k$ concepts, downsampled to $16$ px & $+0.045$ \\
    Control (re-embed unchanged)               & $\approx 0$ \\
    \bottomrule
  \end{tabular}
\end{table}

\end{document}